\documentclass{article}
\makeatletter
\newif\ificmlapptoc
\icmlapptocfalse

\newcommand{\icmlapptocstart}{\global\icmlapptoctrue}

\let\icml@origl@section\l@section
\let\icml@origl@subsection\l@subsection

\def\l@section#1#2{%
  \ificmlapptoc
    \icml@origl@section{#1}{#2}%
  \fi
}

\def\l@subsection#1#2{%
  \ificmlapptoc
    \icml@origl@subsection{#1}{#2}%
  \fi
}

\newcommand{\icmlprintappendixtoc}{%
  \begingroup
  \@starttoc{toc}%
  \endgroup
}
\makeatother

\usepackage{microtype}
\usepackage{graphicx}
\usepackage{subcaption}
\usepackage{booktabs} %
\usepackage{tablefootnote} %
\usepackage{threeparttable}
\newcounter{sharedtabfn}

\usepackage{hyperref}

\usepackage[preprint]{icml2026}

\usepackage{amsmath}
\usepackage{amssymb}
\usepackage{mathtools}
\usepackage{amsthm}
\usepackage{algorithm}
\usepackage{algorithmic}
\usepackage{amsfonts,bm}
\usepackage{tikz}
\usepackage{bbm}
\usepackage{pgfplots}
\usepackage{tablefootnote}
\usepackage{pgfplotstable}
\pgfplotsset{compat=1.18} %
\usepackage{booktabs} 
\usepackage{multirow}
\usepackage[labelfont=bf]{caption}
\usepackage{nicefrac}
\usepackage{graphicx}
\usepackage{subcaption}
\usepackage{tcolorbox}

\usepackage{url}
\newcommand{\panel}[1]{\includegraphics[width=\linewidth]{#1}}
\usepackage[capitalize,noabbrev]{cleveref}

\def\eqref#1{equation~\ref{#1}}

\def\1{\bm{1}}

\DeclareMathAlphabet{\mathsfit}{\encodingdefault}{\sfdefault}{m}{sl}
\SetMathAlphabet{\mathsfit}{bold}{\encodingdefault}{\sfdefault}{bx}{n}

\DeclareMathOperator{\sign}{sign}

\newtheorem{theorem}{Theorem}

\newtheorem{remark}{Remark}
\theoremstyle{definition}
\newtheorem{definition}{Definition}
\crefname{definition}{Definition}{Definitions}
\Crefname{definition}{Definition}{Definitions}
\newtheorem{assumption}{Assumption}
\crefname{assumption}{Assumption}{Assumptions}
\Crefname{assumption}{Assumption}{Assumptions}

\newcommand{\x}{\mathbf{x}}
\newcommand{\xk}{\x_k}
\newcommand{\xkone}{\x_{k+1}}
\newcommand{\xpareto}{\x^\dagger}
\newcommand{\statespace}{\mathcal{S}}
\newcommand{\cost}{\ell}
\newcommand{\costi}{\cost^{i}}
\newcommand{\xstari}{\x^{*,i}}

\newcommand{\bb}[1]{\mathbb{#1}}
\newcommand{\bargaininggame}[2]{\mathcal{B}_{#1}(#2)}
\newcommand{\gradcosti}[1]{\nabla_{#1}\costi}
\newcommand{\dibs}{\texttt{DiBS}}
\newcommand{\dibsmtl}{\texttt{DiBS-MTL}}
\newcommand{\ball}[2]{\mathcal{V}(#1, #2)}
\newcommand{\normalizedgradi}[2]{\frac{\gradcosti{#1}(#2)}{\|\gradcosti{#1}(#2)\|_2}}
\newcommand{\distoptimali}[1]{\|#1-\xstari\|_2}
\newcommand{\RETURN}{\STATE \textbf{return} }

\usepackage[textsize=tiny]{todonotes}

\icmltitlerunning{Monotonic Transformation Invariant Multi-task Learning}

\begin{document}

\twocolumn[
  \icmltitle{Monotonic Transformation Invariant Multi-task Learning}

  \icmlsetsymbol{equal}{*}

  \begin{icmlauthorlist}
    \icmlauthor{Surya Murthy}{equal,yyy}
    \icmlauthor{Kushagra Gupta}{equal,yyy}
    \icmlauthor{Mustafa Karabag}{yyy}
    \icmlauthor{David Fridovich-Keil}{yyy}
    \icmlauthor{Ufuk Topcu}{yyy}
  \end{icmlauthorlist}

  \icmlaffiliation{yyy}{The University of Texas at Austin, TX 78712, USA}

  \icmlcorrespondingauthor{Surya Murthy}{surya.murthy@utexas.edu}
  \icmlcorrespondingauthor{Kushagra Gupta}{kushagrag@utexas.edu}

  \icmlkeywords{Machine Learning, ICML}

  \vskip 0.3in
]

\printAffiliationsAndNotice{}  %

\begin{abstract}
    Multi-task learning (MTL) algorithms typically rely on schemes that combine different task losses or their gradients through weighted averaging. These methods aim to find Pareto stationary points by using heuristics that require access to task loss values, gradients, or both. In doing so, a central challenge arises because task losses can be arbitrarily scaled relative to one another, causing certain tasks to dominate training and degrade overall performance. A recent advance in cooperative bargaining theory, the Direction-based Bargaining Solution (\dibs), yields Pareto stationary solutions immune to task domination because of its invariance to monotonic nonaffine task loss transformations. However, the convergence behavior of \dibs~in nonconvex MTL settings is currently not understood. To this end, we prove that under standard assumptions, a subsequence of \dibs~iterates converges to a Pareto stationary point when task losses are nonconvex, and propose \texttt{DiBS-MTL}, an adaptation of \dibs~to the MTL setting which is more computationally efficient that prior bargaining-inspired MTL approaches. Finally, we empirically show that \texttt{DiBS-MTL} is competitive with leading MTL methods on standard benchmarks, and it drastically outperforms state-of-the-art baselines in multiple examples with poorly-scaled task losses, highlighting the importance of invariance to nonaffine monotonic transformations of the loss landscape. Code available at \url{https://github.com/suryakmurthy/dibs-mtl}
\end{abstract}

\section{Introduction}\label{section: intro}
 The successes of deep learning have inspired investigation into ``generalist" networks---models simultaneously trained for learning multiple tasks. As a result, numerous multi-task learning (MTL) algorithms have been developed to tackle the inevitable conflict between task-specific loss gradients, aiming to ensure that during training, no task is under-optimized compared to others \citep{kendall2018multi, sener2018multi, yu2020gradient, liu2021conflict, navon2022multi, liu2023famo}.
However, most existing MTL methods are not robust against nonaffine (monotonic) transformations to task losses, which is a crucial property desirable in the context of deep learning---where same preferences can be represented with losses of different nonaffine scalings, and it is unclear which relative scaling of the different losses ensures balanced learning without expensive and exhaustive ablations. We consider the problem of multi-task learning (MTL) through the lens of centralized, cooperative bargaining methods that are invariant to \emph{nonaffine} monotonic task loss transformations.

The issue of different task losses being directly incomparable and scaled in different, nonaffine fashions arises very naturally in many deep learning domains. For instance, reinforcement learning applications demand that a practitioner leverages prior, task-specific domain knowledge to design an effective reward function \citep{yu2025reward}. However, in a downstream MTL setting, the loss corresponding to this reward function may dominate (or get dominated by) other task losses. At the same time, the relative performance of a ``good" task-specific policy does not change when the corresponding task loss is monotonically transformed, i.e., the transformation does not change the actual underlying preferences over options. \emph{Thus, in an MTL problem, the available task losses can be seen as monotonic, possibly nonaffine, transformations of some underlying set of ideal, unknown task losses that are meaningfully scaled with respect to each other.}

Recent work in MTL has developed a connection with cooperative game theory \citep{navon2022multi}. In this setting, each different loss function is a separate agent in a bargaining game, and the idea is to find a balanced Pareto optimum among the agents' objectives.
Classical solutions to these bargaining games (e.g., Nash, as explored by \citet{navon2022multi}) are \textit{not} robust to nonaffine monotonic scalings, and only recently has a technique---Direction-based Bargaining Solution (\texttt{DiBS})---been developed which remains \textit{invariant} to these transformations \citep{gupta2025cooperative}.
However, the convergence of \texttt{DiBS} has only been analyzed in settings with strongly convex losses, and it is unclear to what extent the favorable properties of \texttt{DiBS} will apply in realistic MTL applications where task losses are almost always nonconvex. Inspired by this, we investigate the following:
\begin{enumerate}
    \item \emph{What theoretical properties can be established for Direction-based Bargaining Solution (\texttt{DiBS}) in the general setting where agent objectives (task losses) can be \underline{nonconvex}?}
    
    \textbf{Contribution 1.} We show that under standard assumptions, for nonconvex losses, a subsequence of the \texttt{DiBS} iterates converges to a Pareto stationary point asymptotically. Notably, our result does not require the linear independence of task loss gradients at non-Pareto stationary points, an assumption that is required by MTL methods using the Nash bargaining solution to deliver the same asymptotic guarantee \citep{navon2022multi}.

    \item \emph{Can \texttt{DiBS} readily adapt to MTL applications? If so, how does it compare to existing MTL methods?}

    \textbf{Contribution 2.} We extend \texttt{DiBS} to multi-task learning, showing its natural compatibility with existing bargaining-for-learning frameworks. To this end, we propose \dibsmtl, which adapts \texttt{DiBS} to MTL utilizing local loss approximations, is computationally more efficient than previous bargaining-inspired MTL approaches, and also preserves desirable invariance to nonaffine monotonic task loss transformations. We empirically show that \texttt{DiBS-MTL} performs competitively with state-of-the-art MTL methods on widely used vision and quantum chemistry MTL benchmarks.

    \textbf{Contribution 3.} We further investigate settings where task losses are nonaffinely and poorly scaled relative to each other in two domains---a challenging synthetic problem that exhibits substantial nonconvexity (and which highlights Contribution 1), and a larger-scale multi-task reinforcement learning benchmark. We show that \dibsmtl~drastically outperforms state-of-the-art baselines in these domains; highlighting the importance of MTL methods to be invariant to nonaffine monotonic transformations of the loss landscape.

\end{enumerate}

\section{On related MTL works and existing bargaining solutions}\label{section: related work}
\subsection{Related MTL literature}
The most popular MTL approach in practice is linear scalarization (\texttt{LS})---constructing a scalarized loss by taking the unweighted sum of task losses, or using known static coefficients to compute a weighted average. While previous work has advocated for \texttt{LS} methods \citep{kurin2022defense, xin2022current}, in practice, \texttt{LS} can lead to situations where certain tasks remain under-optimized. Further, it has been shown that \texttt{LS} also does not necessarily recover the entire Pareto front generated by the task losses \citep{hu2023revisiting}. Other methods tackle the MTL problem through more sophisticated multiobjective optimization approaches, aiming to find Pareto optimal (or stationary, in general nonconvex settings) points.  Such methods seek to address the task imbalances arising during training via heuristics that (i) use task-specific loss values to compute a scalar weighted average loss (but with evolving weights, unlike \texttt{LS}) \citep{kendall2018multi, liu2019end, lin2022reasonable, liu2023famo}, or (ii) use task-specific loss gradients to find update directions iteratively during training \citep{sener2018multi, yu2020gradient, chen2020just, liu2021conflict, liu2021towards, navon2022multi}.

Existing loss-based heuristics include maximizing improvement of the worst-performing task \citep{liu2023famo}, forcing improvements to be similar across tasks \citep{liu2019end}, weighting task losses randomly \citep{lin2022reasonable}, weighting task losses according to relative improvements \citep{chen2018gradnorm}, and adapting weights according to task-based uncertainty measures  \citep{kendall2018multi}. Heuristics for gradient-based approaches include finding mutual task improvement directions \citep{yu2020gradient}, probabilistically masking task gradients according to their sign \citep{chen2020just}, computing weights that minimize the norm of the convex combination of task gradients \citep{sener2018multi}, using gradient-based approaches to maximize improvement of the worst-performing task \citep{liu2021conflict}, and finding the Nash bargaining solution for a bargaining subproblem  \citep{navon2022multi}. However, as we empirically demonstrate in \Cref{section: experiments}, these \emph{existing methods are \underline{not} robust to monotonic, nonaffine task loss transformations}, potentially experiencing significant performance drops in such settings. 
We remark than one gradient-based method, \texttt{IMTL-G} \citep{liu2021towards}, in principle can produce solutions invariant to monotonic nonaffine task loss transformations. However, we show in Appendix \ref{appendix: imtlg} that even for simple convex problems, \texttt{IMTL-G} can converge to Pareto solutions that heavily favor one task over the other.

Finally, we note that other approaches also exist for the MTL problem, such as (i) task clustering, where methods first cluster tasks to reduce conflicts, and then update model training parameters for each cluster \citep{standley2020tasks, fifty2021efficiently, song2022efficient, shen2024go4align}; and (ii) parameter sharing methods, which design neural network architectures consisting of task-specific and task-shared modules/parameters \citep{kokkinos2017ubernet, guo2020learning, gao2020mtl}.
These approaches differ from the proposed approach at a fundamental level, in that they try to design a suite of models for the space of tasks, or design MTL model architectures, whereas the proposed approach aims to optimize \textit{a single} model (with a pre-defined architecture), balancing the performance of all tasks.

\subsection{Preliminaries on cooperative bargaining games}
We provide a brief background of cooperative bargaining theory, which we utilize in our main contribution. A thorough description can be found in classical literature \citep{thomson1994cooperative, narahari2014game}. A centralized bargaining game consists of $N$ agents and a mediator; and $\x\in\statespace\subseteq\bb{R}^n$ denotes the state of the game. We assume the $i^{\textrm{th}}$ agent has a differentiable cost $\costi(\x):\statespace\to\bb{R}, i\in[N]$. In the context of MTL, every task can be thought of as being represented by one agent, with the agent's cost being the task loss.

Each agent want's to minimize its cost, and has preferred states $\xstari\in\arg\min_{\x\in\statespace}\costi(\x)$ it wants the game to go towards. For nonconvex costs, this could be a local minimum. The goal of the mediator is to execute a bargaining strategy and find a solution state $\x^\dagger$. 
Let $\bm{\cost}(\x)=[\cost^i(\x),\dots,\cost^N(\x)]$. For convenience, we denote such a bargaining game by $\bargaininggame{\statespace}{\bm{\cost}}$. 
Numerous bargaining solutions have been proposed in economics literature \citep{thomson1994cooperative}, each employing different heuristics to find a Pareto optimal point---a point at which no update direction exists, which simultaneous decreases the losses for all agents. In the case when the agent losses $\costi(\x)$ are nonconvex, a more appropriate goal for the mediator is to find a \emph{Pareto stationary} point, which is a first-order necessary condition for Pareto optimality.

\begin{definition}[\textbf{Pareto stationarity}] \label{def: pareto stationarity}
    For $\bargaininggame{\statespace}{\bm{\cost}}$, a point $\x^\dagger\in\statespace$ is Pareto stationary if $\exists~\beta^i\geq 0, i\in[N]$, such that $\sum_{i\in[N]}\beta^i\gradcosti{\x}(\x^\dagger) = 0$, and $\sum_{i\in[N]}\beta^i=1$.
\end{definition}
We now highlight the Direction-based Bargaining Solution (\dibs), which we use in our method. For a bargaining game $\bargaininggame{\statespace}{\bm\cost}$, \dibs~finds a Pareto stationary point by taking an initial point $\x_1\in\statespace$ and running the iterations
\begin{align}\label{eq: dibs iterations}
    \nonumber &\xkone = \xk - h(\xk)\\:=&\xk - \alpha_k\cdot\sum_{i\in[N]}\left(\distoptimali{\xk}\cdot\normalizedgradi{\x}{\xk}\right).
\end{align}
\begin{remark}[\textbf{Invariance of} \dibs]
    For monotonic, possibly nonaffine transformations $h^1,\dots,h^N$, let $\bm{h}(\bm\cost)(\x)=[h^1(\cost^i)(\x),\dots,h^N(\cost^N)(\x)]$. Then \dibs~produces the same solution for the bargaining games $\bargaininggame{\statespace}{\bm{h}(\bm \cost)}$ and $\bargaininggame{\statespace}{\bm \cost}$, for the same initial point $\x_1\in\statespace$ and sequence of stepsizes $\{\alpha_k\}_{k\geq 0}$. This invariance to monotonic nonaffine transformations is because \dibs~only utilizes normalized gradients and locally preferred states, i.e., local minima, both of which do not change under such transformations.
\end{remark}
We also note that prior has work has used the classical Nash bargaining solution (\texttt{NBS}) \citep{nash1950bargaining} in the context of MTL \citep{navon2022multi}. However, the \texttt{NBS} is invariant to only affine agent (or task) loss transformations, and can change if the agent (or task) losses undergo monotonic \emph{nonaffine} transformations.

As highlighted in \Cref{section: intro}, \emph{the need for MTL methods which are robust to monotonic nonaffine transformations makes it natural to consider incorporating \dibs---which is invariant to such distortions---in an MTL approach}. However, since practical MTL problems typically involve nonconvex losses, it is crucial to examine the behavior of \dibs~in the nonconvex regime, a setting not yet analyzed in prior work. This serves as our motivation for the next section.

\section{What can we say about \dibs~in the nonconvex regime?}
In this section we establish a convergence guarantee for \dibs~in the setting when agent losses can be nonconvex, which is often the case in practical MTL problems. Currently, guarantees only exist for the case when all agents have strongly convex losses, under which \dibs~enjoys global asymptotic convergence to a Pareto stationary point \citep{gupta2025cooperative}. We begin by highlighting our assumptions.
\begin{assumption}\label{assumption: well posed}
    For $\bargaininggame{\statespace}{\bm\cost}$, the set of Pareto stationary points lies in the interior of $\statespace$, and all $\xstari$ exist, are finite, and are also in the interior of $\statespace$. The agent costs $\costi$ are differentiable, and bounded below.
\end{assumption}

\begin{assumption}[\textbf{Relaxed in \Cref{section: relaxing boundedness assumption}}] \label{assumption: bounded}
    The sequence $\{\x_k\}_{k=1}^{\infty}$ generated by the \dibs~iterations given in \Cref{eq: dibs iterations} are bounded, i.e., there is a bounded set $\mathcal{D}\subseteq\statespace$ such that $\xk\in\mathcal{D}~\forall~k\in\bb{N}$. 
\end{assumption}
\Cref{assumption: well posed} is standard and ensures that the problem is well posed. \Cref{assumption: bounded} is a temporary assumption that we make for a clear exposition of our arguments while studying convergence of \dibs~in nonconvex settings. \Cref{assumption: bounded} makes the \dibs~iterates bounded, and we relax this assumption in \Cref{section: relaxing boundedness assumption}, showing that standard techniques from dynamical systems theory can be used to ensure \dibs~iterates are bounded, forgoing the need for \Cref{assumption: bounded}.

We now present our main theorem, the proof of which is in Appendix \ref{appendix: proof of thm}.
\begin{theorem}\label{thm: subsequence converges to pareto}
    Let $\{\xk\}_{k=1}^{\infty}$ be the sequence generated by the \dibs~algorithm given in \Cref{eq: dibs iterations}, for an initial point $\x_1\in\statespace$ and stepsizes that follow the Robbins-Monro conditions, i.e., $\sum_k \alpha_k = \infty$, $\sum_k \alpha_k^2 < \infty$ \citep{robbins1951stochastic}. Then, under \Cref{assumption: well posed,assumption: bounded}, the sequence $\{\xk\}_{k=1}^{\infty}$ has a subsequence that asymptotically converges to a Pareto stationary point $\x^\dagger$, i.e., $h(\xpareto)=0$.
\end{theorem}
\Cref{thm: subsequence converges to pareto} establishes that the sequence produced by the \dibs~iterates has a subsequence which converges to a Pareto stationary point, even when the agent losses are nonconvex. Note that this result is similar to a guarantee presented for the Nash bargaining solution in prior work \citep{navon2022multi}. However, the existing result in prior work requires the assumption that agents' loss gradients are linearly independent at all non-Pareto stationary points. It is unclear to what extent realistic MTL problems satisfy this assumption. In comparison, our result does not require such a linear independence assumption.

\begin{algorithm}[ht]
\caption{$T$-step \dibsmtl}
\label{alg:dibs-mtl}
\begin{algorithmic}

\REQUIRE Initial parameters $\theta_{0}$, losses $\{\ell^i\}_{i=1}^N$, 
learning rate $\eta$, approximation radius $\epsilon$, step size $\alpha < \nicefrac{\epsilon}{N}$, 
number of epochs $J$

\FOR{$j = 1$ to $J$}

    \STATE Compute normalized gradients 
    $\bar{g}^i_{j-1}= \normalizedgradi{\theta}{\theta_{j-1}}$

    \STATE For each $i$, compute preferred individual parameters inside  $\epsilon$-radius:    $\theta^{*,i}_{j} = \theta_{j-1} - \epsilon \bar{g}^i_{j-1},
    $

    \STATE Set $\theta_{(j-1,0)} \leftarrow \theta_{j-1}$

    \STATE \texttt{// DiBS updates inside approximation radius $\epsilon$}

    \FOR{$t = 1$ to $T$}

        \STATE Compute \texttt{DiBS} update direction:
        
        $d_{(j,t)} =
        \sum_{i\in[N]}
        \Big(
        \|\theta_{(j-1,t-1)} - \theta^{*, i}_{j} \|_2
        \cdot
        \bar{g}^i_{j-1}
        \Big)$

        \STATE Update:
        $\theta_{(j-1,t)} \leftarrow
        \theta_{(j-1,t-1)} - \alpha\, d_{(j,t)}$

    \ENDFOR

    \STATE $\Delta \theta_{j-1} =
    \theta_{(j-1,T)} - \theta_{(j-1,0)}$

    \STATE Update:
    $\theta_{j} \leftarrow
    \theta_{j-1} + \eta\, \Delta \theta_{j-1}$

\ENDFOR

\RETURN $\theta_{j}$

\end{algorithmic}
\end{algorithm}
\section{Adapting \dibs~to multi-task learning}
We now extend \dibs~to the multi-task learning setting. We begin by introducing some notation, and then describe the underlying bargaining game for MTL, which has been proposed in previous work \citep{navon2022multi}.
\paragraph{Notation.} We use $\bm 0$ to denote a zero vector of appropriate dimensions, and $\ball{\x}{r}$ denotes a ball of radius $r$, centered at $\x$. 
The vector $\theta\in\bb{R}^n$ represents shared task parameters; any additional task-specific parameters are suppressed for notational convenience and do not influence the calculation of $\theta$ iterates in this section. $j$ denotes an epoch iteration, whereas $t$ will denote bargaining updates \emph{within} an epoch.
\paragraph{Bargaining for MTL.} The goal in MTL is to train a model to perform multiple tasks. Every learning task has an associated task loss $\costi(\theta):\bb{R}^n\to\bb{R}$. During training, given parameters $\theta_{j}$ at some epoch $j+1$, bargaining is iteratively conducted during the epoch to find the next set of parameters $\theta_{j+1} = \theta_{j} + \eta \Delta\theta_{j}$ for some learning rate $\eta$. 

To this end, every task is represented as a distinct bargaining agent with the (minimization) objective being an approximation of the difference $\costi(\theta_{j}+\eta \Delta\theta_{j}) - \costi(\theta_{j})$. We consider a first-order approximation, in line with prior bargaining work for a fair comparison \citep{navon2022multi}, with the $i^\text{th}$ agent's objective being
\begin{align*}
    \min_{\|\Delta\theta_{j}\|_2\leq\epsilon} \underbrace{\left(\gradcosti{\theta}(\theta_{j})\right)^\top\Delta\theta_{j}}_{:=\omega^i(\Delta\theta_{j})},
\end{align*}
where we constrain the update vector $\Delta\theta_{j}$ to lie in $\ball{\bm 0}{\epsilon}$, to prevent overshooting caused by large update steps. \emph{Thus, for an MTL problem with $N$ tasks---the bargaining game becomes $\bargaininggame{\ball{\bm 0}{\epsilon}}{[\omega^1,\dots,\omega^N]}$}.

\paragraph{\dibs~for MTL.} We now proceed to apply \dibs~to the bargaining game $\bargaininggame{\ball{\bm 0}{\epsilon}}{[\omega^1,\dots,\omega^N]}$. Note that the bargaining game has linear objectives, and thus the normalized gradient
\begin{align*}
    \frac{\nabla_{\Delta\theta_{j}}\omega^i}{\|\nabla_{\Delta\theta_{j}}\omega^i\|_2} = \frac{\gradcosti{\theta}(\theta_{j})}{\|\gradcosti{\theta}(\theta_{j})\|_2} := \bar{g}^i_j
\end{align*}
only needs to be computed once at the starting of a bargaining game. Further, the distance of the individual optima from the initial point in \Cref{eq: dibs iterations} has a closed form solution such that $\Delta \theta^{*,i}_{j} :=  \theta^{*,i}_{j} - \theta_{j-1} =-\left(\nicefrac{\gradcosti{\theta}(\theta_{j})}{\|\gradcosti{\theta}(\theta_{j})\|_2}\right)\cdot\epsilon$ for the linear objectives, i.e., an individual optimum is the furthest point in the negative gradient direction, given that $\Delta\theta_{j}$ is constrained to lie in $\ball{\bm 0}{\epsilon}$. Thus, for an initial choice $\Delta\theta_{(j,0)} = \bm 0$ and stepsize sequence $\{\alpha_t\}$, $T$-iterate \dibs~ for MTL satisfies
\begin{align}\label{eq: multi step dibs}
    \nonumber&\Delta\theta_{(j, t+1)} = \Delta\theta_{(j,t)}
    - \alpha_t\cdot\sum_{i\in[N]}\Bigg(\left\|\Delta\theta_{(j,t)}+\epsilon\cdot\bar{g}^i_j\right\|_2\cdot
    \bar{g}^i_j \Bigg) \tag{\texttt{T-step DiBS-MTL}}
\end{align}
Using this the parameters between epochs are updated as in \Cref{alg:dibs-mtl}. In practice, single-step versions of such iterative schemes are common \citep{liu2023famo}. Such a version described below still \emph{preserves} the desirable invariance to monotonic, nonaffine task loss transformations:
\begin{align}\label{eq: dibs mtl}
    \Delta\theta_{(j,1)} = - \epsilon\cdot\sum_{i\in[N]}\bar{g}^i_j\tag{\texttt{1-step DiBS-MTL}} 
\end{align}

We remark that unlike multi-step variants, \ref{eq: dibs mtl} \emph{does not} have explicit dependencies on the individual optima $\theta^{*,i}_{j}$ due to the first-order approximation combined with the ball constraint. Further, \texttt{1-step DiBS-MTL} does not introduce any additional hyperparameters as the $\epsilon$ term is absorbed into the learning rate $\eta$. In \Cref{sec:toy,subsection: cv/mtrl,section: additional results} we provide results showing that the single-step version performs similarly to the multi-step version despite being relatively simpler on a variety of MTL problems.

\definecolor{mplblue}{HTML}{1F77B4}
\definecolor{mplorange}{HTML}{FF7F0E}
\definecolor{mplgreen}{HTML}{2CA02C}
\definecolor{mplred}{HTML}{D62728}
\definecolor{mpgrey}{HTML}{7F7F7F}
\definecolor{mplpurple}{RGB}{148,103,189}
\definecolor{mplbrown}{RGB}{140,86,75}
\definecolor{mplpink}{RGB}{227,119,194}
\newcommand{\legendline}[2]{\textcolor{#1}{\rule[0.6ex]{14pt}{2pt}}\hspace{0.4em}#2}

\section{Experiments}\label{section: experiments}
We now evaluate \dibsmtl~in standard MTL benchmarks extensively used in literature \citep{yu2020gradient, liu2021conflict, navon2022multi, liu2023famo}, a demonstrative two objective example, a multi-task reinforcement learning benchmark (Meta-World \texttt{MT10}) \citep{yu2020meta}, and supervised learning computer vision and graph neural network benchmarks (\texttt{NYU-v2}, \texttt{Cityscapes} and \texttt{QM9}) \citep{silberman2012indoor, cordts2016cityscapes, blum2009970}. The results for the \texttt{Cityscapes} benchmark are included in Appendix \ref{appendix: cityscapes}. \textbf{Our main aims are: (i)} to verify that \dibsmtl~ consistently converges to Pareto-optimal solutions from diverse initializations in nonconvex settings, \textbf{(ii)} to show that due to robustness against monotonic nonaffine task loss transformations, \dibsmtl~significantly outperforms state-of-the-art MTL baselines in examples where task losses are poorly scaled relative to each other, and \textbf{(iii)} to confirm that leveraging normalized gradients to achieve this robustness does not compromise the solution quality of \dibsmtl~ and that \dibsmtl~is competitive to state-of-the-art baselines across large-scale MTL benchmarks.
\begin{figure}[t]
  \centering

  \includegraphics[width=0.8\linewidth]{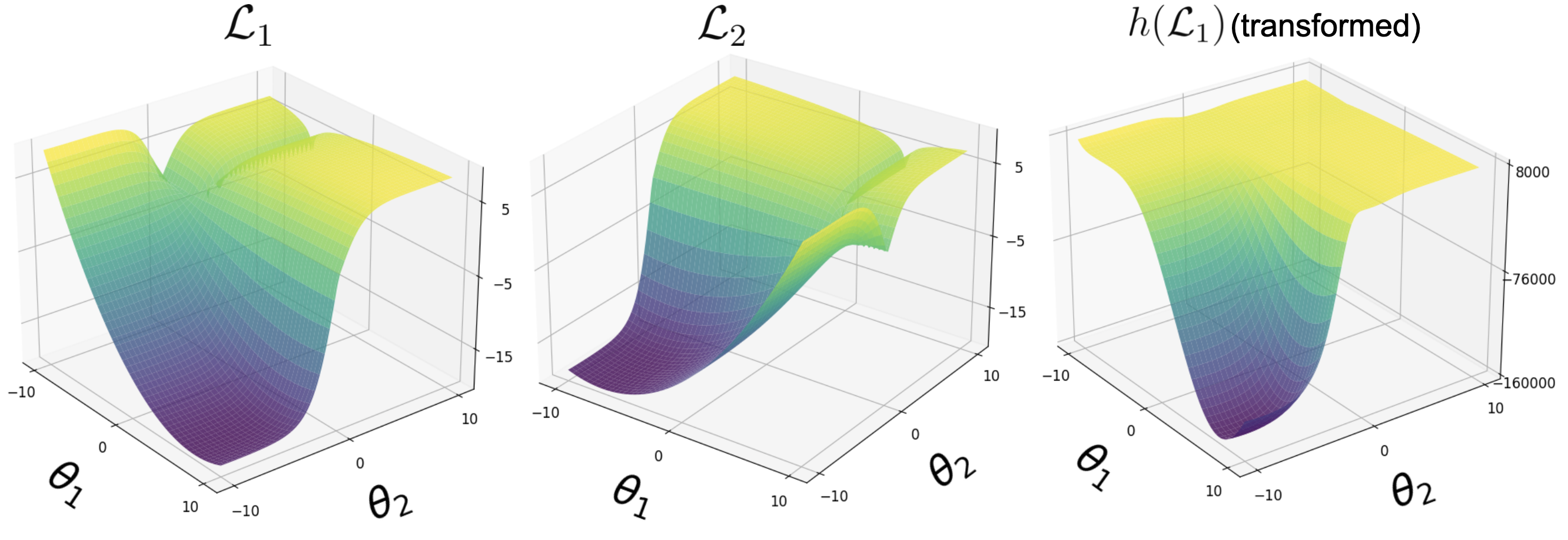}

  \caption{Loss functions in the nonconvex example (\Cref{sec:toy}). We employ the nonaffine transformation $h(\mathcal{L}_1) = \sign(\mathcal{L}_1)\cdot \mathcal{L}_1^4$.}
  \label{fig:loss_curves}
\end{figure}

\begin{figure*}[ht]
  \centering

  \includegraphics[width=0.9\linewidth]{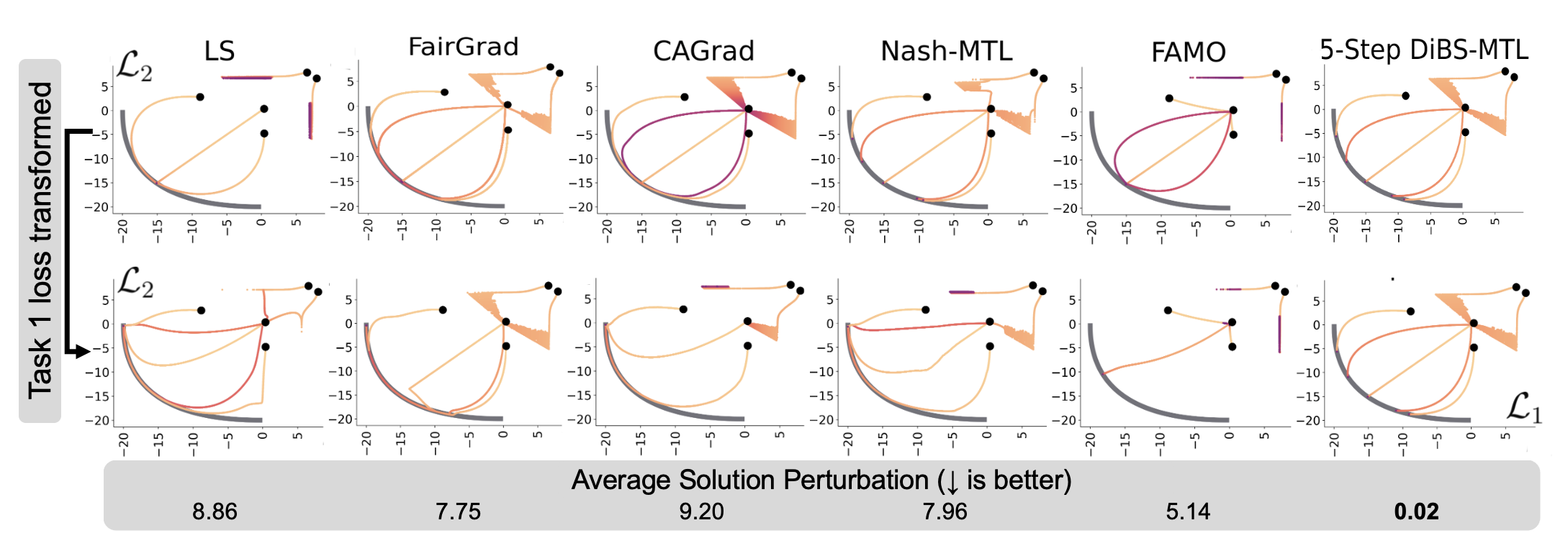}

  \caption{\dibsmtl~successfully avoids task domination due to its invariance to monotonic nonaffine transformations, while baseline MTL methods visibly favor task 1 when $\mathcal{L}_1$ is transformed, and some baseline methods even fail to reach the Pareto front. $\bullet$ and \legendline{mpgrey}{} denote initializations and the Pareto front respectively. To better visualize results for the transformed case, we retain the original $\mathcal{L}_1$ axis. Results for multi-step \dibsmtl~and additional baselines are included in \Cref{fig:extra_trajectories}.}
  \label{fig:trajectories}
\end{figure*}

\subsection{Does \dibsmtl~consistently converge to Pareto solutions from diverse initializations?} \label{sec:toy}

\paragraph{Setting.} We first test $\dibsmtl$ on a challenging two-dimensional, nonconvex multi-objective optimization example with two objectives $(\mathcal{L}_1, \mathcal{L}_2)$, which is a standard benchmark for illustrating the ability of MTL methods to achieve balanced Pareto solutions \citep{yu2020gradient, liu2021conflict, navon2022multi, liu2023famo, ban2024fair}. As shown in \Cref{fig:loss_curves}, each objective (provided in Appendix~\ref{app:toy-eqs}) in this example has deep valleys, with the bottoms having a large magnitude difference between the objective gradients, and a high (positive) curvature.  Such phenomenon is documented to exist when optimizing neural networks as well, and can lead to one objective (task) dominating others \citep{goodfellow2014qualitatively, yu2020gradient}.
\paragraph{Baselines.} We test \texttt{5-step} \dibsmtl~along with 5 baselines---(1) linear scalarization (\texttt{LS}), which optimizes the unweighted loss average, (2) \texttt{CAGrad} \citep{liu2021conflict}, (3) \texttt{FairGrad} \citep{ban2024fair}, (4) \texttt{Nash-MTL} \citep{navon2022multi}, and (5) \texttt{FAMO}\footnote{We were unable to reproduce the baseline plots reported in the original FAMO paper \citep{liu2023famo}.  The results shown here were obtained by strictly following the default parameters and instructions in the publicly available \texttt{FAMO} repository \url{https://github.com/Cranial-XIX/FAMO}.} \citep{liu2023famo}.
\paragraph{\dibsmtl~yields Pareto solutions from diverse initializations, and unlike other methods, \dibsmtl~is invariant to monotonic nonaffine objective transformations.} \Cref{fig:trajectories} plots the solutions generated by \dibsmtl~for different initializations, for the costs $(\mathcal{L}_1, \mathcal{L}_2)$ above, as well as for $\left(h\left(\mathcal{L}_1\right), \mathcal{L}_2\right)$ where we apply the monotonic nonaffine transformation $h(x)=\sign(x)\cdot x^4$ to $\mathcal{L}_1$. We observe that \emph{$\dibsmtl$ consistently leads to balanced Pareto solutions for all initializations}, and these solutions remain invariant to the transformation. It is also evident that unlike \dibsmtl, all MTL baselines produce significantly biased solutions once $\mathcal{L}_1$ is transformed (i.e., comparing the top and bottom rows of \Cref{fig:trajectories}), signaling that \emph{baselines are susceptible to task domination} when tasks are differently scaled. Further, \dibsmtl~achieves faster runtimes than the previous bargaining-inspired MTL method \texttt{Nash-MTL} (see \Cref{subsection: runtime}).

\begin{table*}[t]
  \centering
  \caption{\texttt{1-step DiBS-MTL} achieves robustness to the different reward transformations on the \texttt{MT10} multi-task reinforcement learning benchmark, and significantly outperforms all MTL baselines for all transformed settings. We report best-checkpoint success (evaluated every 200 episodes) mean results $\pm$ standard error over 10 seeds. \textbf{Bold} values indicate best performances per column.\\ $^\dag$ For certain transformations, training with UW, Nash-MTL, FairGrad, and CAGrad was highly sensitive to loss scaling. This instability caused the training process to diverge, leading the underlying MuJoCo simulator \citep{todorov2012mujoco} to repeatedly produce out-of-bounds errors. As a result, we were unable to collect results for these cases.}
  \label{tab:mt10-all-combined-reversed}

  \setlength{\tabcolsep}{6pt}

  \begin{threeparttable}
    \fontsize{8.5pt}{10pt}\selectfont
    \begin{tabular}{lcccc}
      \toprule
      & \multicolumn{4}{c}{Transformed task} \\   
      \cmidrule(lr){2-5}
      \addlinespace[0.3em]
       & No task transformed & \texttt{reach} transformed & \texttt{window open} transformed & \texttt{peg insert} transformed \\
      \midrule      
      \texttt{FAMO} & $0.920 \pm 0.020$ & $0.700 \pm 0.075$ & $0.830 \pm 0.033$ & $0.570 \pm 0.070$ \\
      \texttt{LS} & $0.850 \pm 0.040$ & $0.480 \pm 0.044$ & $0.200 \pm 0.026$ & $0.230 \pm 0.030$ \\
      \texttt{UW} & $0.830 \pm 0.033$ & $0.650 \pm 0.081$ & ---$^\dag$ & $0.420 \pm 0.074$ \\
      \texttt{FairGrad} & $0.950 \pm 0.022$ & $0.290 \pm 0.023$ & ---$^\dag$ & ---$^\dag$ \\
      \texttt{Nash-MTL} & $\bm{0.970 \pm 0.015}$ & ---$^\dag$ & ---$^\dag$ & ---$^\dag$ \\
      \texttt{CAGrad} & $0.950 \pm 0.022$ & $0.689 \pm 0.031$ & ---$^\dag$ & ---$^\dag$ \\
      \midrule
      \texttt{1-step DiBS-MTL} & $0.890 \pm 0.023$ & $\bm{0.850 \pm 0.022}$ & $\bm{0.920 \pm 0.025}$ & $\bm{0.700 \pm 0.037}$ \\
      \bottomrule
    \end{tabular}

  \end{threeparttable}
\end{table*}

\begin{figure*}[t]
  \centering

  \legendline{mplblue}{\texttt{1-step} ~\dibsmtl}\hspace{1.2em}
  \legendline{mplorange}{\texttt{FAMO}}\hspace{1.2em}
  \legendline{mplred}{\texttt{UW}}\hspace{1.2em}
  \legendline{mplgreen}{\texttt{LS}}\hspace{1.2em}
  \legendline{mplpurple}{\texttt{FairGrad}}
{\hspace{1.2em}
  \legendline{mplbrown}{\texttt{Nash-MTL}}}
  {\hspace{1.2em}
  \legendline{mplpink}{\texttt{CAGrad}}}

  \vspace{0.4em}

  \begin{subfigure}[t]{0.23\linewidth}
    \centering
    \includegraphics[width=\linewidth]{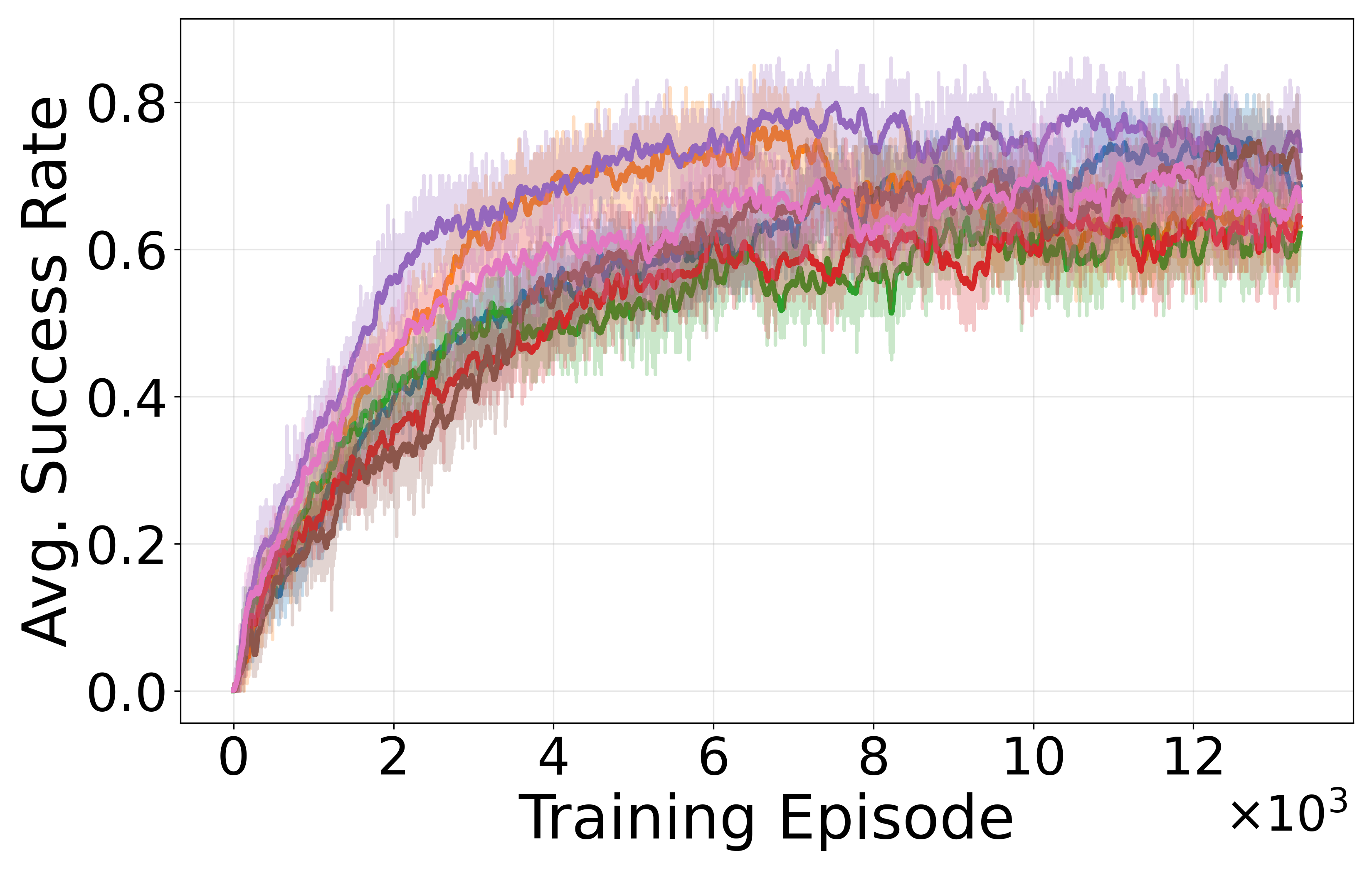}
    \caption{No task transformed}
    \label{fig:mt10-nominal}
  \end{subfigure}\hfill
  \begin{subfigure}[t]{0.23\linewidth}
    \centering
    \includegraphics[width=\linewidth]{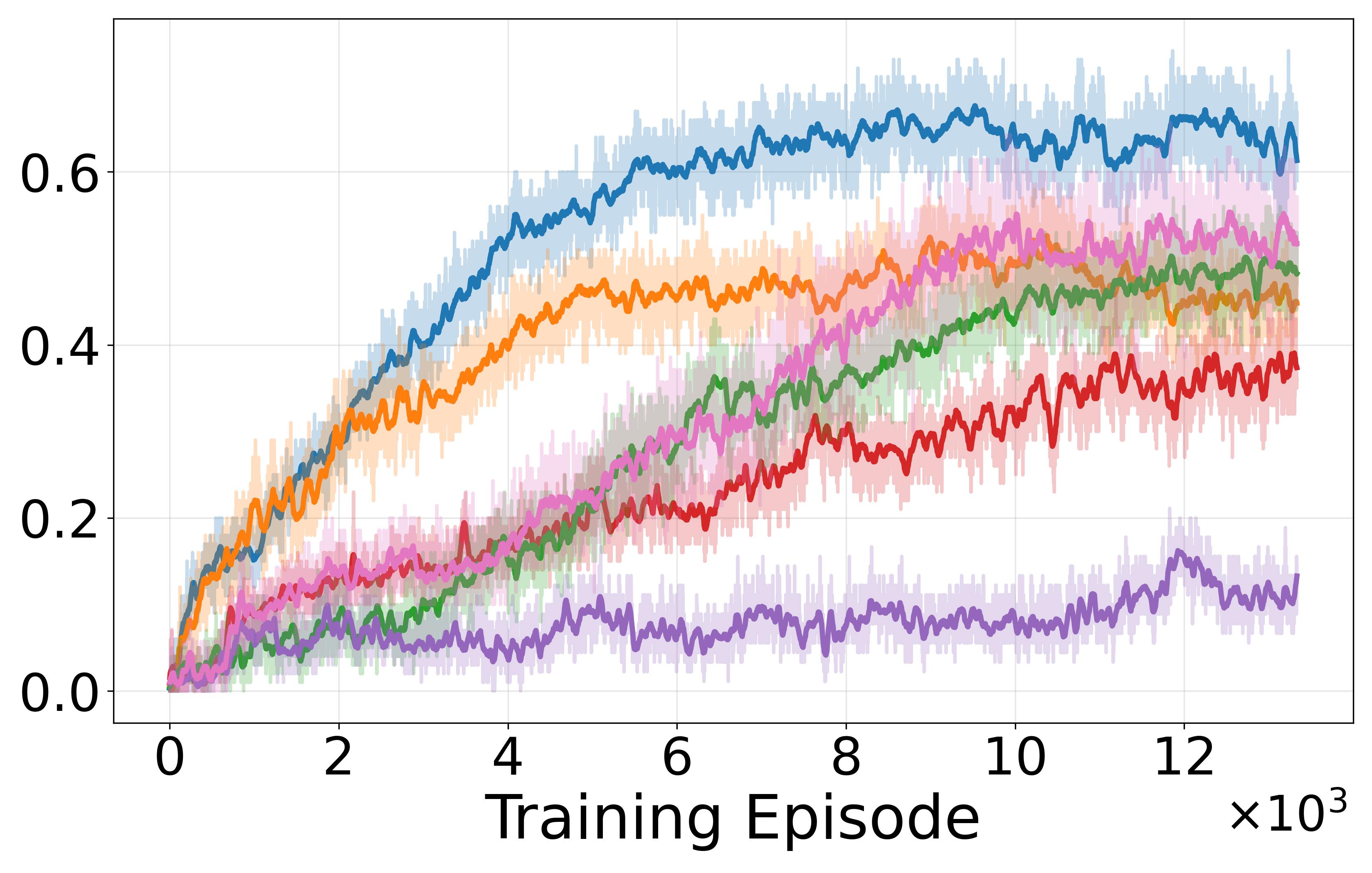}
    \caption{\texttt{reach} transformed}
    \label{fig:mt10-reach-transformed}
  \end{subfigure}\hfill
  \begin{subfigure}[t]{0.23\linewidth}
    \centering
    \includegraphics[width=\linewidth]{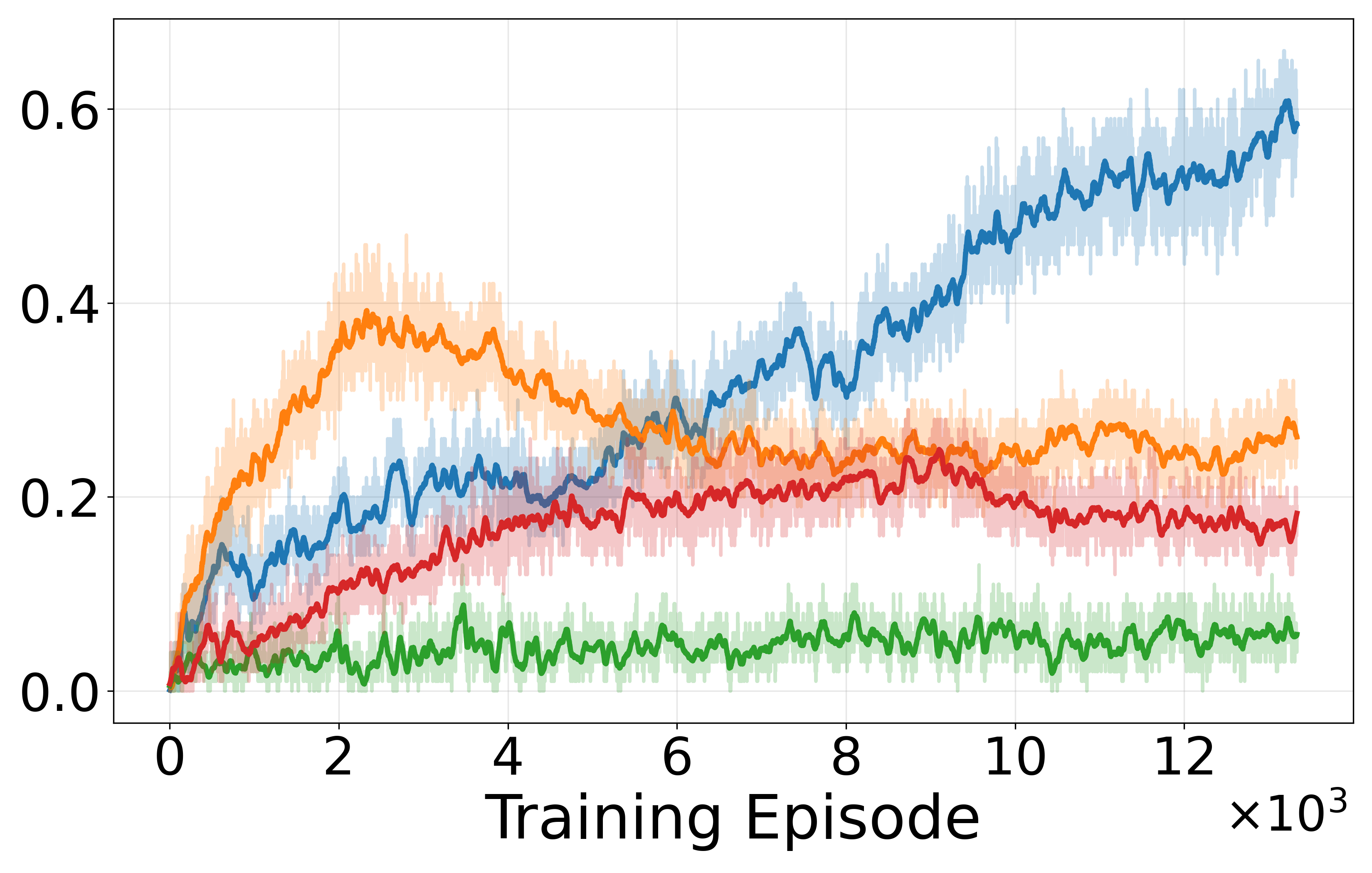}
    \caption{\texttt{peg insert} transformed}
    \label{fig:mt10-peg-transformed}
  \end{subfigure}\hfill
  \begin{subfigure}[t]{0.23\linewidth}
    \centering
    \includegraphics[width=\linewidth]{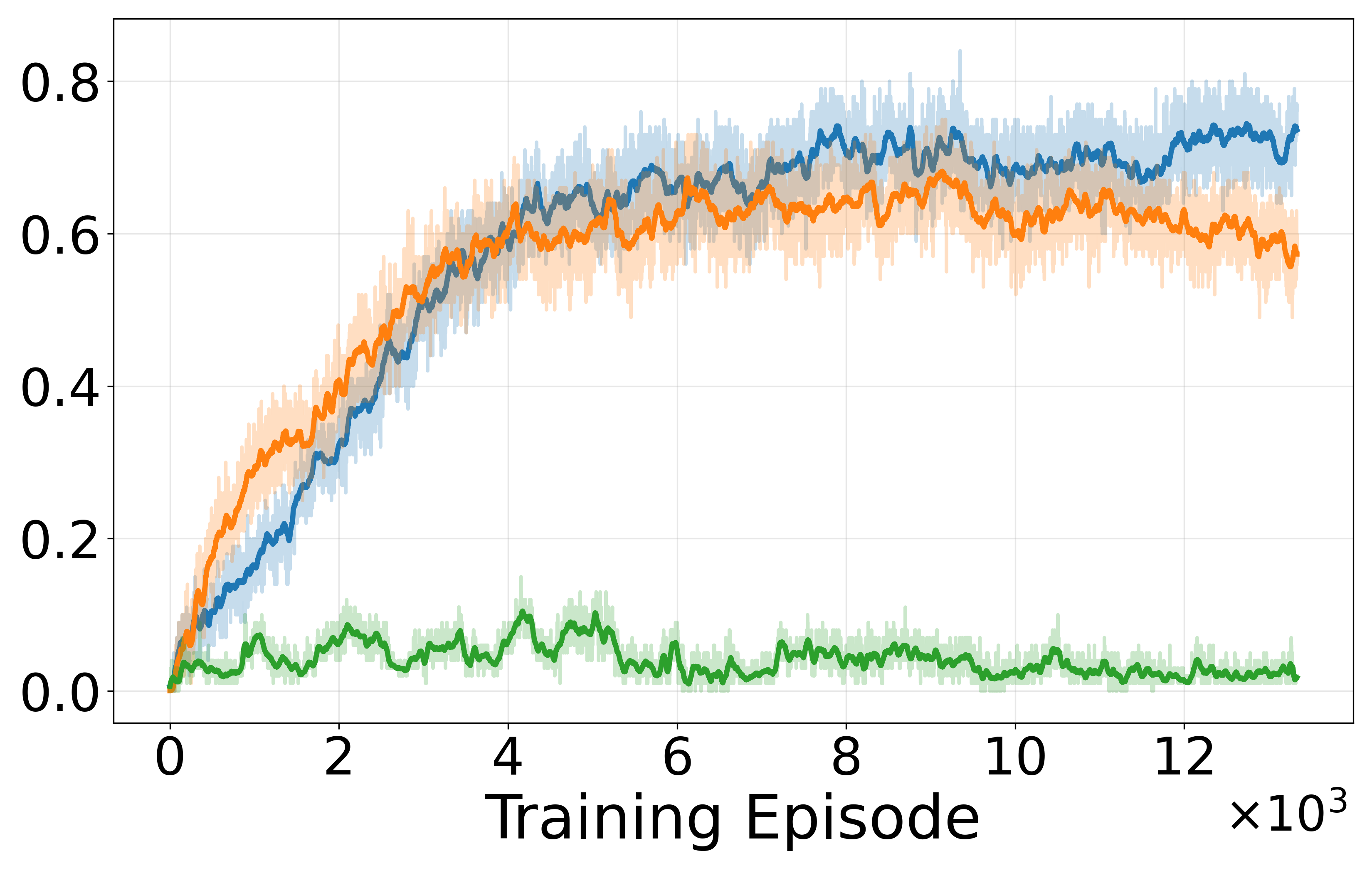}
    \caption{\texttt{window open} transformed}
    \label{fig:mt10-window-transformed}
  \end{subfigure}

  \caption{Average task success rate on the MT10 benchmark, averaged over all tasks and 10 random seeds. Transparent curves show per-episode success rates, while solid curves denote rolling averages computed over a window of 51 episodes. \texttt{1-step DiBS-MTL} demonstrates strong robustness to diverse reward transformations and consistently and significantly outperforms all baselines across all transformed settings.} 
  \label{fig:mt10}
\end{figure*}

\subsection{Does the use of normalized gradients in \texttt{DiBS-MTL} induce robustness to nonaffine transformations?}\label{subsection: mtrl}
We now conduct experiments in the multi-task reinforcement learning (MTRL) Meta-World \texttt{MT10} benchmark \cite{yu2020meta} to investigate how \dibsmtl~and baseline MTL methods perform when some tasks undergo nonaffine transformations. We limit our study in this section to MTRL, because task-specific losses in supervised learning examples such as computer vision are relatively standardized in the literature and thus transformations there will be unwarranted \citep{wang2022comprehensive, azad2023loss}. In comparison, as we mention in \Cref{section: intro}, in the MTRL setting, rewards for different tasks might be designed on very different scales. 

\paragraph{Metaworld\texttt{ MT10} \citep{yu2020meta}.} This benchmark consists of a robot arm conducting 10 tasks in the MuJoCo simulator \citep{todorov2012mujoco}, each with distinct reward functions. The list of the 10 tasks is given in Appendix \ref{appendix: implementation}.

\paragraph{MTRL baselines and metrics.} As all \dibsmtl~variants are transformation invariant, we restrict our study to \ref{eq: dibs mtl}, which has the lowest per epoch complexity. We compare \ref{eq: dibs mtl} with (1) \texttt{LS}, (2) \texttt{UW}, (3) \texttt{FAMO}, (4) \texttt{Fairgrad} \citep{ban2024fair}, (5) \texttt{Nash-MTL} \citep{navon2022multi} and (6) \texttt{CAGrad} \citep{liu2021conflict}. Following prior work, for every method, we use Soft Actor-Critic (SAC) \citep{haarnoja2018soft} as the underlying reinforcement learning algorithm, hyperparameters values are taken from the corresponding baseline works, and we report the fraction of tasks successfully completed ~\citep{navon2022multi, liu2023famo}, where the definition of task success is as described in the Meta-World benchmark \citep{yu2020meta}. Reported results have been averaged over 10 random seeds. Complete implementation details and hyperparameter values are in Appendix \ref{appendix: implementation}.

\paragraph{Task transformations in the MTRL setup.} We consider transformation of 3 different tasks---for each, we apply monotonic, nonaffine transformations to the underlying reward function. For actor-critic algorithms like SAC, this transformation corresponds to potentially nonmonotonic, nonaffine critic loss transformations. This is because the losses fit the collected rewards to value functions rather than fitting the state dependent rewards. Thus, these task reward transformations can lead to potentially \emph{nonmonotonic} task loss transformations. However, we remark that in the policy landscape, a good policy for the nominal reward is also likely to perform well for the transformed reward.

\paragraph{Reward transformations.} We consider the following task-transformation pairs: (i) $h(r) = \sign(r)\cdot r^4$ for \texttt{reach}, (ii) $h(r) =(5+r)^4$ for \texttt{peg insert}, ($h$ is monotonic over the reward range), and (iii) $h(r)=\exp(r)$ for \texttt{window open}. These transformations were arbitrarily selected to illustrate a range of functional forms and tasks.

\paragraph{\dibsmtl~displays robustness against nonaffine reward transformations, while baselines drastically degrade.} \Cref{fig:mt10-nominal,fig:mt10-reach-transformed,fig:mt10-peg-transformed,fig:mt10-window-transformed} and \Cref{tab:mt10-all-combined-reversed} show that \ref{eq: dibs mtl} achieves remarkable robustness to the different reward transformations compared to existing MTL methods. While gradient-based MTL methods like \texttt{FairGrad}, \texttt{Nash-MTL} and \texttt{CAGrad} display strong performance in the nominal setting, their performance significantly degrades in the transformed cases, often to the extent that training became unstable, and the underlying MuJoCo simulator \citep{todorov2012mujoco} consistently gave out of bound errors for many seeds. Out of all methods, \ref{eq: dibs mtl} \textit{achieves state-of-the-art overall robustness for all transformed cases}, while other methods either fail or exhibit significant performance drops for all transformations compared to their performance in the nominal setting. Finally, \dibsmtl~again achieves faster runtimes than the previous bargaining-inspired MTL method \texttt{Nash-MTL} (see \Cref{subsection: runtime}).
\begin{tcolorbox}[leftrule=1.5mm,top=1mm,bottom=0mm]
\begin{center}
    \textbf{Takeaway from \Cref{sec:toy,subsection: mtrl}:}
\end{center}
\dibsmtl~significantly outperforms state-of-the-art MTL baselines in all examples where underlying task losses/rewards are poorly scaled relative to each other, thanks to its monotonic transformation invariance. This highlights the unique suitability of \dibsmtl~to MTL problems with heterogeneous, possibly unknown task loss scalings.
\end{tcolorbox}
\begin{table*}[t]
  \centering
  \caption{\dibsmtl~variants exhibit strong performance on the \texttt{NYU-v2} benchmark, achieving state of the art values on MTL relevant metrics ($\Delta m\%$ \& MR). \textbf{Bold}/\underline{underline} indicates best/second-best performance per column. Results averaged over 3 random seeds.}
  \label{tab:nyu}
  \resizebox{\textwidth}{!}{
  \begingroup
  \setlength{\tabcolsep}{6pt}
  \begin{tabular}{@{}lccccccccccc@{}}
    \toprule
    & \multicolumn{2}{c}{Segmentation} 
    & \multicolumn{2}{c}{Depth Estimation} 
    & \multicolumn{5}{c}{Surface Normal} & & \\
    \cmidrule(lr){2-3} \cmidrule(lr){4-5} \cmidrule(lr){6-10}
    & 
    mIoU $\uparrow$ & 
    Pix Acc $\uparrow$ & 
    Abs Err $\downarrow$ & 
    Rel Err $\downarrow$ & 
    Median $\downarrow$ & Mean $\downarrow$ & $<30$ $\uparrow$ & $<22.5$ $\uparrow$ & $<11.25$ $\uparrow$ &
    $\Delta m\%$ $\downarrow$ & MR $\downarrow$\\
    \midrule

    \texttt{STL} &
    38.30 & 63.76 &
    0.68 & 0.28 &
    19.21 & 25.01 & 69.15 & 57.20 & 30.14 &
    --- & --- \\
    \midrule

    \texttt{MGDA} &
    32.03 & 60.77 &
    0.6102 & 0.2453 &
    \textbf{19.00} & \textbf{24.64} & \textbf{69.83} & \textbf{57.78} & \textbf{30.55} &
    -0.69 & 11.0 \\

    \texttt{UW} &
    39.08 & 64.73 &
    0.5464 & 0.2284 &
    23.04 & 27.34 & 62.85 & 49.23 & 23.49 &
    3.77 & 8.2 \\

    \texttt{Nash-MTL} &
    40.16 & 65.65 &
    \underline{0.5331} & 0.2203 &
    19.96 & 25.25 & 68.29 & 55.72 & 28.62 &
    -3.98 & 4.2 \\

    \texttt{FAMO} &
    37.58 & 64.08 &
    0.5595 & 0.2296 &
    \underline{19.15} & \underline{25.04} & \underline{69.36} & \underline{57.44} & \underline{30.23} &
    -3.82 & 8.5 \\

    \texttt{LS} &
    40.16 & 65.63 &
    0.5445 & 0.2223 &
    23.03 & 27.50 & 62.66 & 49.32 & 23.61 &
    3.06 & 6.0 \\

    \texttt{GO4Align} &
    38.87 & 64.52 &
    0.5698 & 0.2406 &
    23.41 & 27.81 & 61.83 & 48.56 & 23.45 &
    5.47 & 9.9 \\

    \texttt{IGBv2} &
    33.61 & 61.88 &
    0.5728 & 0.2222 &
    23.18 & 27.94 & 62.14 & 49.10 & 23.59 &
    11.92 & 9.7 \\

    \texttt{IGBv2+Nash} &
    38.81 & 65.66 &
    \textbf{0.5258} & \textbf{0.2086} &
    20.09 & 25.50 & 67.82 & 55.35 & 28.37 &
    1.67 & 5.1 \\

    \texttt{LDC} &
    39.21 & 65.70 &
    0.5337 & \underline{0.2180} &
    20.23 & 25.29 & 67.98 & 55.19 & 27.98 &
    -3.24 & 4.5 \\
    
    \midrule
    \texttt{1-step DiBS-MTL} &
    40.92 & 66.60 &
    0.5337 & 0.2216 &
    20.06 & 25.35 & 68.15 & 55.54 & 28.54 &
    -4.11 & \textbf{3.1} \\

    \texttt{5-step DiBS-MTL} &
    \underline{41.27} & \textbf{67.10} &
    0.5473 & 0.2300 &
    19.69 & 25.13 & 68.69 & 56.28 & 29.31 &
    \underline{-4.57} & 4.2 \\

    \texttt{10-step DiBS-MTL} &
    \textbf{41.33} & \underline{67.09} &
    0.5381 & 0.2260 &
    19.77 & 25.06 & 68.67 & 56.12 & 29.02 &
    \textbf{-4.74} & \underline{3.6} \\

    \bottomrule
  \end{tabular}
  \endgroup
  }
\end{table*}

\begin{table*}[t!]
    \centering
    \caption{\dibsmtl~exhibits strong performance on the \texttt{QM9} benchmark, with the 10-step variant achieving best and second-best values on MTL relevant metrics $\Delta m\%$ \& MR. \textbf{Bold}/\underline{underline} indicates best/second-best performance. Results averaged over 3 random seeds.}
    \resizebox{\textwidth}{!}{%
    \begin{tabular}{lrrrrrrrrrrrrr}
    \toprule
     & $\mu$ & $\alpha$ & $\epsilon_\text{HOMO}$ & $\epsilon_\text{LUMO}$ & $\langle R^2\rangle$ & ZPVE & $U_0$ & $U$ & $H$ & $G$ & $c_v$ & \multirow{2}{*}{$\Delta m\% \downarrow$} & \multirow{2}{*}{MR $\downarrow$} \\
      \cmidrule(lr){2-12}
      & \multicolumn{11}{c}{MAE $\downarrow$} & & \\
    \midrule 
        \texttt{STL}      & 0.07 & 0.18 & 60.60 & 53.90 & 0.50 & 4.53 & 58.80 & 64.20 & 63.80 & 66.20 & 0.07 & --- & --- \\
    \midrule
    
        \texttt{LS}       & \textbf{0.0998} & 0.3108 & \textbf{69.83} & \textbf{84.87} & 5.13 & 13.43 & 137.25 & 137.97 & 138.43 & 133.55 & 0.1226 & 167.16 & 7.42 \\
        
        \texttt{GO4ALIGN}  & 0.1047 & 0.3175 & \underline{70.40} & 86.48 & 5.05 & 13.35 & 138.78 & 139.49 & 139.86 & 135.27 & 0.1252 & 167.90 & 7.88 \\
        
        \texttt{UW}       & 0.4123 & 0.3937 & 159.07 & 145.41 & \textbf{1.09} & 4.85 & \underline{61.52} & \underline{61.95} & \underline{61.98} & \underline{61.40} & 0.1206 & 102.96 & 5.50 \\
        
        \texttt{LDC}      & 0.2782 & 0.3318 & 141.53 & 121.37 & 1.20 & \textbf{4.34} & \textbf{51.43} & \textbf{51.83} & \textbf{51.91} & \textbf{51.28} & 0.1076 & \underline{68.79} & \textbf{3.50} \\
        
        \texttt{FAMO}     & 0.4422 & 0.3998 & 158.41 & 145.83 & \underline{1.10} & 5.04 & 65.76 & 66.15 & 66.24 & 65.40 & 0.1267 & 110.99 & 6.67 \\
        
        \texttt{Nash-MTL}  & \underline{0.1024} & 0.3153 & 70.75 & \underline{85.29} & 5.05 & 13.04 & 135.67 & 136.55 & 136.82 & 132.71 & 0.1227 & 165.12 & 6.96 \\
        
         \texttt{IGBV2}   & 0.2866 & 0.3557 & 149.29 & 135.04 & 1.24 & \underline{4.60} & 64.68 & 65.08 & 65.15 & 64.92 & 0.1176 & 85.75 & 5.33 \\
        
        \texttt{IGBV2 + Nash} &  0.4399 & 0.3762 & 142.19 & 139.73 & 1.62 & 5.10 & 67.20 & 67.61 & 67.66 & 67.13 & 0.1159 & 115.24 & 7.17 \\
        
        \midrule
        \texttt{1-step DiBS-MTL}  & 0.1347 & \textbf{0.2841} & 97.53 & 95.01 & 2.56 & 5.75 & 65.92 & 66.38 & 66.39 & 65.83 & 0.1067 & 71.83 & \underline{4.83} \\
        
        \texttt{5-step DiBS-MTL} & 0.1285 & 0.2914 & 96.40 & 95.95 & 2.55 & 5.68 & 69.28 & 69.68 & 69.74 & 69.11 & \underline{0.1065} & 72.92 & 5.54 \\
        
        \texttt{10-step DiBS-MTL} & 0.1213 & \underline{0.2860} & 93.50 & 90.46 & 2.37 & 5.81 & 70.12 & 70.54 & 70.75 & 69.83 & \textbf{0.1061} & \textbf{67.37} & 5.21 \\
        
    \bottomrule 
    \end{tabular}
    }
    \label{tab:qm9}
\end{table*}

\subsection{Can the transformation robust DiBS-MTL maintain competitive performance in untransformed settings on standard MTL benchmarks? }
\label{subsection: cv/mtrl}
To compare \dibsmtl~with existing methods in large-scale learning examples, we perform experiments in supervised learning in two domains---(i) computer vision (the \texttt{NYU-v2} dataset \citep{silberman2012indoor}), and (ii) graph neural networks for quantum chemistry (\texttt{QM9 dataset} \citep{blum2009970}). Additional results for the computer vision \texttt{Cityscapes} dataset \citep{cordts2016cityscapes} are included in \Cref{appendix: cityscapes}. All of these are widely used challenging benchmarks in MTL literature \citep{liu2021conflict, navon2022multi, liu2023famo, xiao2025ldc, dai2023improvable, shen2024go4align, ban2024fair}, and complete benchmark descriptions are given in \Cref{appendix: implementation}. We begin by describing the baselines and metrics for all datasets. We keep the default training procedures established in prior baseline works for a fair comparison. All implementation details are given in Appendix \ref{appendix: implementation}.

\paragraph{Supervised MTL baselines.}
For all datasets, we compare \texttt{1-Step}, \texttt{5-Step}, and \texttt{10-Step} \dibsmtl\ with an extensive suite of strong MTL baselines: (1) linear scalarization (\texttt{LS}), (2) uncertainty weighting (\texttt{UW}) \citep{kendall2018multi}, (3) \texttt{MGDA} \citep{sener2018multi}, (4) \texttt{Nash-MTL} \citep{navon2022multi}, (5) \texttt{FAMO} \citep{liu2023famo}, (6) \texttt{GO4Align} \citep{shen2024go4align}, (7) \texttt{IGBv2}, (8) \texttt{IGBv2+Nash} \citep{dai2023improvable}, and (9) \texttt{LDC} \citep{xiao2025ldc}. For all baselines across all datasets, the architecture, data pipeline, protocols, and training hyperparameters are taken to be the same as that in the corresponding works introducing the baselines.

\paragraph{Supervised MTL metrics.}
In addition to task-specific metrics, we evaluate all methods using two standard measures of multi-task performance: (1) the mean relative per-task performance drop $\Delta m\%$ with respect to single-task learning (\texttt{STL}), and (2) the mean rank (\texttt{MR}) across all task metrics. $\Delta m\%$ captures the average improvement or degradation relative to independent single task training (STL), while \texttt{MR} summarizes the average rank of a method across all tasks for a benchmark. These metrics are used in all baseline works as well, to assess whether an MTL method improves joint learning across diverse tasks.

\paragraph{\dibsmtl\ achieves competitive performance across all datasets.}
Table~\ref{tab:nyu} reports results on \texttt{NYU-v2}. \dibsmtl~variants \textit{achieve state-of-the-art performance in $\Delta m\%$ as well as \texttt{MR}}, being the best or second-best in both MTL relevant metrics across all methods, demonstrating balanced improvements across segmentation, depth, and normal prediction. \dibsmtl~variants also achieve competitive or superior performance in segmentation and depth estimation, while matching strong baselines on surface normal prediction. In the \texttt{QM9} benchmark, \texttt{DiBS-MTL} variants display superior performance for the $\alpha$ and $c_v$ prediction tasks. Additionally, \texttt{DiBS-MTL} variants \textit{achieve the best $\Delta m\%$ and second best \texttt{MR} performance} (\Cref{tab:qm9}). Finally, \ref{eq: dibs mtl} is faster than \texttt{Nash-MTL} for \texttt{NYU-v2}, and only slightly slower on \texttt{QM9} (\Cref{subsection: runtime}).
\begin{tcolorbox}[leftrule=1.5mm,top=1mm,bottom=0mm]
\begin{center}
    \textbf{Takeaway from \Cref{subsection: cv/mtrl}:}
\end{center}
Across all large-scale untransformed benchmark evaluations, \dibsmtl~ consistently performs competitively against leading MTL baselines. This indicates that the transformation robustness property of \dibsmtl~does not sacrifice performance on popular real-world benchmarks.
\end{tcolorbox}

\section{Conclusion}
We consider the problem of constructing multi-task learning (MTL) methods that are robust to monotonic, nonaffine task loss transformations. 
In practice, different task losses can be arbitrarily scaled with respect to one another, and those scalings can be understood as monotonic, possibly nonaffine transformations of some ideal, unknown losses which are meaningfully comparable.
To address this problem, we present \dibsmtl, an MTL method which is invariant to such transformations because it uses only normalized task loss gradients. 
Building upon recent work which formalizes the MTL training update step as a bargaining game played between tasks \citep{navon2022multi}, we adapt the recently introduced \texttt{DiBS} bargaining approach \citep{gupta2025cooperative} in order to find a Pareto stationary (and monotonic transformation-invariant) update direction.
While \dibs~enjoys convergence guarantees when losses are strongly convex, practical MTL problems almost always have nonconvex task losses. To bridge this gap, we prove that a subsequence of the \texttt{DiBS} iterates asymptotically converges to a Pareto stationary point even in the nonconvex regime. Empirically, we demonstrate that (i) due to its monotonic transformation invariance, \dibsmtl~significantly outperforms state-of-the-art MTL baselines in settings with task losses being poorly scaled relative to each other, highlighting the unique suitability of \dibsmtl~to MTL problems with heterogeneous, possibly unknown task loss scalings; and (ii) the transformation robustness property of \dibsmtl~does not sacrifice performance on real-world benchmarks, where \dibsmtl~is competitive against leading MTL baselines.

\section*{Impact Statement}

This paper presents work whose goal is to advance the field of Machine
Learning. There are many potential societal consequences of our work, none
which we feel must be specifically highlighted here.
\section*{Acknowledgments}
This work was supported by the Air Force Office for Scientific Research under grant number FA9550-22-1-0403, Lockheed Martin under number MRA16-005-RPP023, DARPA Transfer from Imprecise and Abstract Models to Autonomous Technologies (TIAMAT) under grant number HR0011-24-9-0431, the National Science Foundation under awards 2336840 and 2211548, and by the Army Research Laboratory under Cooperative Agreement Number W911NF-25-2-0021.

\nocite{langley00}

\bibliography{example_paper}
\bibliographystyle{icml2026}

\newpage
\appendix
\onecolumn

\appendix

\addtocontents{toc}{\protect\icmlapptocstart}

\section*{Contents}
\makeatletter
\icmlprintappendixtoc
\makeatother

\section{Proof of \Cref{thm: subsequence converges to pareto}}\label{appendix: proof of thm}
\paragraph{Proof sketch.} As the iterates are bounded, there exists a subsequence which converges to some cluster point $\xpareto$. To show that $\xpareto$ is a Pareto stationary point, we will show by contradiction that $h(\xpareto)=0$.
\begin{proof}
    Because the \dibs~iterates $\{\xk\}_{k=1}^\infty$ is a bounded sequence from \Cref{assumption: bounded} (relaxed in \Cref{section: relaxing boundedness assumption}), by Bolzano-Weierstrass theorem, $\exists~\xpareto\in\statespace$ such that a subsequence of the sequence $\{\xk\}_{k=1}^\infty$ converges to $\xpareto$, and that for every neighborhood $\mathcal{U}$ of $\xpareto$, there exist infinite $n\in\bb{N}$ such that $\x_n\in\mathcal{U}$. We will show by contradiction that $h(\xpareto)=0$. Let there exist an $a >0$ such that $\|h(\xpareto)\|_2=a$. We define
    \begin{align*}
M:=\sup_{\x\in\mathcal{D}}\|h(\x)\|_2, \quad\quad\quad u := \frac{h(\xpareto)}{\|h(\xpareto)\|_2}
    \end{align*}Then, by continuity of $u^\top h(\x)$, we have
    \begin{align}
        \forall~\delta>0, \exists~\epsilon>0 &\text{ such that }\|\x-\xpareto\|_2\leq \delta \implies u^\top h(\x) \geq a\epsilon,  \label{eq: proof ae condition}\\
       \text{ and } \exists~N\in\bb{N} &\text{ such that }k\geq N \implies \alpha_k \leq \frac{a}{C} \text{ for some }C>0. \label{eq: condition on alpha}
    \end{align}
    Let $\ball{\x}{r}$ denote a ball in $\bb{R}^n$ centered at $\x\in\bb{R}^n$, with radius $r$. Let us pick a $\tilde\delta>0$, and analyse the \dibs~iterations when they are in $\ball{\xpareto}{\tilde\delta}$. Because there an infinite number of $n$ such that $\x_n\in \ball{\xpareto}{\tilde\delta}$, there are two possibilities:
    \begin{enumerate}
        \item \textbf{Case 1.} $\exists~\delta>\tilde\delta$ such that the \dibs~iterates enter $\ball{\xpareto}{\tilde\delta}$, exit $\ball{\xpareto}{ \delta}$, and then re-enter $\ball{\xpareto}{\tilde\delta}$---an infinite number of times.
        \item \textbf{Case 2.}  The \dibs~iterates enter $\ball{\xpareto}{\tilde\delta}$ and then exit it at most a finite number of times before eventually remaining in $\ball{\xpareto}{\tilde\delta}$ forever.
    \end{enumerate}
    \paragraph{Case 1.} Consider the counts when the \dibs~iterates are inside $\ball{\xpareto}{\tilde\delta}$, $n<t_1<t_2<\dots$ such that $\x_{t_k}\in\ball{\xpareto}{\tilde\delta}~\forall~k\in\bb{N}$, and let the earliest corresponding counts when the \dibs~iterates are outside $\ball{\xpareto}{ \delta}$ be $e_k :=\min \{t \geq t_k | \x_t \notin \ball{\xpareto}{ \delta}\}$. Then, from \Cref{eq: proof ae condition}, we have
    \begin{align*}
        u^\top h(\x_t)\geq a\epsilon~\forall~t=t_k,t_{k+1},\dots,e_{k}-1.
    \end{align*}
    Now, from the definition of $t_k$ and $e_k$, we have
    \begin{align}
        \|\x_{e_k} - \x_{t_k}\|_2 &\geq \delta - \tilde\delta > 0\notag\\
        \implies \left\|\sum_{t=t_k}^{e_k-1} \left(\x_{t+1} - \x_t\right)\right\|_2 &\geq \delta - \tilde\delta\notag\\
        \implies \sum_{t=t_k}^{e_k-1}\| \x_{t+1} - \x_t\|_2 &\geq \delta - \tilde\delta \tag{triangle inequality}\\
        \implies \sum_{t=t_k}^{e_k-1}\alpha_t = \sum_{t=t_k}^{e_k-1} \frac{\| \x_{t+1} - \x_t\|_2}{\|h(\x_t)\|_2} &\geq \sum_{t=t_k}^{e_k-1} \frac{\| \x_{t+1} - \x_t\|_2}{M}\geq \frac{\left(\delta - \tilde\delta\right)}{M}\notag\\
        \implies u^\top\left(\x_{e_k} - \x_{t_k}\right) = \sum_{t=t_k}^{e_k-1} u^\top \left(\x_{t+1}-\x_t\right) &= - \sum_{t=t_k}^{e_k-1}\alpha_t u^\top h(\x_t)\notag\\
        &\leq -a\epsilon \sum_{t=t_k}^{e_k-1}\alpha_t \tag{from \Cref{eq: proof ae condition}}\notag\\
        \implies u^\top\left(\x_{e_k} - \x_{t_k}\right)&\leq - \frac{a\epsilon\left(\delta - \tilde\delta\right)}{M}\label{eq: ek - tk}.
    \end{align}
    Here, the first equality could be legitimately written as the definition of case 1 implies that $h(\x_t)\neq 0, t = t_k,\dots,e_k-1$, otherwise the iterates would have remained within $\ball{\xpareto}{\delta}$ forever.
    Now, from the definition of case 1, $t_{k+1}$ and $e_k$, we have that the \dibs~iterates cannot remain outside $\ball{\xpareto}{\tilde\delta}$ for an infinite amount of time, and thus $\exists~C'>0$ such that $C' > \sum_{t=e_k}^{t_{k+1}-1}1~\forall~k$. Thus
    \begin{align}
    u^\top\left(\x_{t_{k+1}} - \x_{e_k}\right) &= \sum_{t=e_k}^{t_{k+1}-1}u^\top(\x_{t+1} - \x_{t})\notag\\
        &\leq \sum_{t=e_k}^{t_{k+1}-1}\alpha_t M \tag{Cauchy-Schwarz}\\
        &=M\sum_{t=e_k}^{t_{k+1}-1}\alpha_t \notag\\
         &\leq \frac{aMC'}{C} \tag{from \Cref{eq: condition on alpha}}\\
        \implies u^\top\left(\x_{t_{k+1}} - \x_{e_k}\right)&\leq \frac{aMC'}{C}:=a\gamma\label{eq: tk+1 - ek}.
    \end{align}
    Combining \Cref{eq: ek - tk,eq: tk+1 - ek}, we get
    \begin{align}
        u^\top\left(\x_{t_{k+1}}-\x_{t_k}\right) \leq  a\left(\gamma -\frac{\epsilon\left(\tilde\delta-\delta\right)}{M}\right)\notag\\
        \implies u^\top\x_{t_{k+1}} \leq u^\top\x_{t_{1}}  + k\cdot a\left(\gamma -\frac{\epsilon\left(\tilde\delta-\delta\right)}{M}\right)\label{eq: tk+1}
    \end{align}
    From Cauchy-Schwarz inequality, $\|u^\top\x_{t_{k+1}}\|_2\leq \|\x_{t_{k+1}}\|_2$ is bounded for all $k\in\bb{N}$ as the \dibs~sequence is bounded. However, from \Cref{eq: tk+1}, as $k$ increases $\|u^\top\x_{t_{k+1}}\|_2$ becomes unbounded, which is a \textbf{contradiction}.

    \paragraph{Case 2.} By definition of case 2, $\exists~ t^*<\infty$ such that $\x_t\in\ball{\xpareto}{\delta}~\forall~t\geq t^*$. Thus for $T > t^*$, we have
    \begin{align}
        u^\top \left(\x_{T} - \x_{t^*}\right) &= \sum_{t=t^*}^{T - 1} u^\top \left( \x_{t+1} - \x_t \right)\notag\\
        &= - \sum_{t=t^*}^{T - 1} \alpha_t u^\top h(\x_t)\notag\\
        &\leq -a\epsilon \sum_{t=t^*}^{T - 1} \alpha_t \tag{from \Cref{eq: proof ae condition}}\\
        \implies u^\top \x_T &\leq u^\top\x_{t^*} - a\epsilon \sum_{t=t^*}^{T - 1} \alpha_t \label{eq: x T}
    \end{align}
    Similar to the argument of contradiction in case 1, $\|u^\top \x_{t^*}\|_2, \|u^\top \x_T\|_2, T\geq t^*$ are bounded. However, \Cref{eq: x T} suggests that $\|u^\top\x_T\|$ becomes unbounded as $T$ increases, because of the Robbins-Monro stepsize condition $\sum_k\alpha_k = \infty$, and $\sum_{k=1}^{t^*-1}\alpha_k$ is finite. Hence, we arrive at a \textbf{contradiction.}

    Thus, from both cases, we get that $h(\xpareto) = 0$, and from \Cref{def: pareto stationarity}, $\xpareto$ is a Pareto stationary point with convex coefficients
    \begin{align*}
        \beta^i = \frac{\nicefrac{\|\xpareto-\xstari\|_2}{\|\gradcosti{\x}(\xpareto)\|_2}}{\sum_{i}\nicefrac{\|\xpareto-\xstari\|_2}{\|\gradcosti{\x}(\xpareto)\|_2}}.
    \end{align*}
\end{proof}

\section{Relaxing Assumption \ref{assumption: bounded}}\label{section: relaxing boundedness assumption}
In the case when all agent (task) losses $\costi$ are convex, it has been established that the \dibs~iterates are bounded \citep{gupta2025cooperative}. In the non-convex setting, while the boundedness of the \dibs~iterates intuitively holds for a problem with gradient conflict, i.e., task loss gradients point in opposite directions, we show that boundedness can be formally guaranteed with standard concepts from dynamical systems theory.

In particular, the following simple modification to \dibs~ allows us to guarantee boundedness, without changing the fact that the only fixed points of the dynamics are Pareto stationary points.
\begin{align*}
    \xkone = \begin{cases}
        f_1(\xk):=\xk - h(\xk),& \text{ if }\|\xk\|_2\leq R \text{ (standard \dibs)}\\
        f_2(\xk):=\xk - \alpha \frac{\xk}{\|\xk\|_2},& \text{ if }\|\xk\|_2\geq R+r \text{ (radially attractive)}\\
        f(\xk, t_k),& \text{ if }R<\|\xk\|_2< R+r \text{ (convex combination)}
    \end{cases}
\end{align*}
\begin{align*}
    \text{where } f(\xk)=\left(1-g(\xk)\right)f_1(\xk) &+ g(\xk)f_2(\xk) + z(\xk, t_k),\\
    z(\xk, t_k) = \left(\frac{\|\xk\|_2-R}{r}\right)\cdot&\left(1-\frac{\|\xk\|_2-R}{r}\right)\cdot\sin(t)\bm{a}\\
    g(\x) = \frac{e^{-\frac{r}{\|x\|_2-R}}}{e^{-\frac{r}{\|x\|_2-R}} + e^{-\frac{1}{1 - \frac{\|x\|_2-R}{r}}}},& \text{ and }
    t_k = \sum_{i=1}^k \mathbbm{1}\{R<\|\x_i\|_2<R+r\}
\end{align*}
Here, $\bm a\in\bb{R}^n$ is any constant vector of the same dimension as the iterates $\xk$. The radius $R$ can be taken to be some large positive number, larger that $\max_{i}\|\xstari\|_2$, and $r$ can be any positive constant. This modification ensures that (i) near the Pareto front, the \dibs~iterates act as usual, (ii) in case the optimization landscape is such that due to a lack of conflicting gradient nature, the iterates somehow move away from the Pareto region and outside the $\|\x\|_2\leq R+r$ ball, the iterates switch to the radially attractive dynamics and return to the ball and remain bounded, (iii) the dynamics switch is smoothly carried out in between the $\|\x\|_2\leq R$ and $\|\x\|_2 \leq R+r$ balls, and (iv) the fixed points of the dynamics do not change--- the time varying term $z(\xk,t_k)$ ensures that only Pareto stationary points (all inside the $\|\x\|_R$ ball) are the fixed points of the modified dynamics, and the dynamics do not converge to non-Pareto stationary points.

\section{Discussion on \texttt{IMTL-G}} \label{appendix: imtlg}
We elaborate on our claim made in \Cref{section: related work} that though \texttt{IMTL-G} \citep{liu2021towards} can be invariant to monotonic nonaffine transformations, it can produce Pareto solutions that are heavily favor one task over the other, possibly leading to task domination issues.

\texttt{IMTL-G} is a gradient-based method, which tries to find an update direction which has an equal projection on all task gradient vectors. Though this equal projection property brings invariance to monotonic nonaffine transformations, it also renders \texttt{IMTL-G} susceptible to task domination, as illustrated by the following simple two dimensional example.

Let $\statespace = [-1, 1]\times[-1, 1]$, with two tasks $\ell^1(x,y) = x^2 + (y-1)^2$ and $\ell^2(x,y) = x^2 + (y+1)^2$.
Then any point of the form $(0,y), y\in[-1,1]$ is a valid solution that \texttt{IMTL-G} can give. Even though all such points are Pareto stationary solutions, only $(0,0)$ is balanced in the sense that it is equidistant to these symmetric functions (one function is a reflection of the other with respect to the X-axis). \emph{Thus \emph{IMTL-G} can output a point like $(0, 0.9)$  which is heavily biased towards task 1.} In contrast, even if the iteratations are started at $(0,0.9)$, \dibs~will return the balanced point $(0,0)$ as a solution.

\section{Additional Supervised MTL Experiments -- \texttt{Cityscapes} benchmark} \label{appendix: cityscapes}

\begin{table*}[t]
  \centering
  \caption{Results for supervised learning \texttt{Cityscapes} benchmark. Segmentation metrics (mIoU, Pix Acc) are shown in percentage form. \textbf{Bold}, \underline{underlined} indicate best and second-best per column.}
  \label{tab:cityscapes}
  \begingroup
  \setlength{\tabcolsep}{6pt}
  \begin{tabular}{@{}lcccccc@{}}
    \toprule
    & \multicolumn{2}{c}{Segmentation (\%)} 
    & \multicolumn{2}{c}{Depth Estimation} 
    &  \\
    \cmidrule(lr){2-3} \cmidrule(lr){4-5} 
    & 
    mIoU $\uparrow$ & 
    Pix Acc $\uparrow$ & 
    Abs Err $\downarrow$ & 
    Rel Err $\downarrow$ &
    $\Delta m\%$ $\downarrow$ &
    MR $\downarrow$
    \\
    \midrule

\texttt{STL}                      & 74.01 & 93.16 & 0.0125 & 27.77 &  ---  & ---  \\
    \midrule

\texttt{MGDA}                     & 69.44 & 91.49 & 0.0137 & 45.26 & 20.03 & 6.3  \\

\texttt{UW}                     & 66.18 & 90.28 & 0.0158 & 54.92 & 34.41 & 10.2  \\

\texttt{Nash-MTL}                 & 73.57 & 92.92 & 0.0140 & 52.34 & 25.30 & 6.0  \\
\texttt{FAMO}                     & 69.80 & 91.66 & 0.0137 & \underline{43.66} & \underline{18.51} & 5.5 \\
\texttt{LS}                       & 71.89 & 92.44 & 0.0180 & 111.89 & 87.64 & 10.0 \\
\texttt{GO4ALIGN}                 & 72.38 & 92.60 & 0.0312 & 93.50 & 97.26 & 9.8 \\
\texttt{IGBV2}                    & 66.44 & 90.48 & 0.0153 & 56.56 & 34.85 & 10.0 \\
\texttt{IGBV2+Nash}               & 73.53 & 92.95 & 0.0135 & 50.42 & 22.58 & 4.5 \\
\texttt{LDC}                      & 75.04 & \underline{93.44} & 0.0144 & \textbf{38.44} & \textbf{12.93} & \textbf{3.0} \\
\midrule
\texttt{DiBS-MTL}                 & 74.64 & 93.31 & 0.0135 & 58.09 & 29.10 & 5.9 \\
\texttt{DiBS-MTL (5 Steps)}       & \textbf{75.46} & \textbf{93.55} & \textbf{0.0132} & 54.88 & 25.27 & \underline{3.2} \\
\texttt{DiBS-MTL (10 Steps)}      & \underline{75.05} & 93.41 & \underline{0.0133} & 52.36 & 23.34 & 3.6 \\

    \bottomrule
  \end{tabular}
  \endgroup
\end{table*}

The \texttt{Cityscapes} dataset \citep{cordts2016cityscapes} contains 5000 street-view $128\times256$, RGB images annotated for two tasks: (1) semantic segmentation with 19 classes and (2) discretized depth estimation with 7 ordinal bins. Following prior MTL evaluations on Cityscapes \citep{navon2022multi, liu2023famo}, we report mIoU and pixel accuracy for segmentation, and absolute and relative depth error for the depth task. This dataset complements \texttt{NYU-v2} by providing a larger-scale, outdoor benchmark with visually and semantically different task relationships.

 \paragraph{Results on \texttt{Cityscapes}.} In the \texttt{Cityscapes} benchmark, \dibsmtl~variants similarly perform competitively against all baselines, achieving strong segmentation performance and favorable depth metrics across the two tasks (see Table~\ref{tab:cityscapes}), and the 5 step variant of \texttt{DiBS-MTL} \textit{achieves the second best 
\texttt{MR}}.

\section{Experimental Details and Code base}\label{appendix: implementation}

\subsection{Demonstrative Nonconvex Example} \label{app:toy-eqs}
\paragraph{Loss Functions.} The non-convex objectives used in $(\mathcal{L}_1, \mathcal{L}_2)$ are given by 
\begin{align*}
\mathcal{L}_1(\theta)=c_{1}(\theta)f_{1}(\theta)+c_{2}(\theta)g_{1}(\theta),&\quad \mathcal{L}_2(\theta)=c_{1}(\theta)f_{2}(\theta)+c_{2}(\theta)g_{2}(\theta)\\
c_{1}(\theta) = \max\bigl(\tanh(0.5\,\theta_{2}),0\bigr),& \quad 
c_{2}(\theta) = \max\bigl(\tanh(-0.5\,\theta_{2}),0\bigr),\\
f_{1}(\theta) = \log \Bigl(\max\bigl(\lvert 0.5(-\theta_{1}-7)&-\tanh(-\theta_{2})\rvert,10^{-6}\bigr)\Bigr) + 6,\\
f_{2}(\theta) = \log \Bigl(\max\bigl(\lvert 0.5(-\theta_{1}+3)&-\tanh(-\theta_{2})+2\rvert,10^{-6}\bigr)\Bigr) + 6,\\
g_{1}(\theta) = \frac{(-\theta_{1}+7)^{2}+0.1(-\theta_{2}-8)^{2}}{10}-20,&\quad
g_{2}(\theta) = \frac{(-\theta_{1}-7)^{2}+0.1(-\theta_{2}-8)^{2}}{10}-20.
\end{align*}
As mentioned in \Cref{section: experiments}, the transformation used on $\mathcal{L}_1$ for the transformed case is the monotonic, nonaffine transformation $h(\ell) = \sign(\ell)\cdot\ell^4$.
\subsection{NYuV2}

\paragraph{Benchmark and Setup.} The NYU-v2 indoor scene dataset~\cite{silberman2012indoor} provides 1,449 RGB-D images with dense per-pixel annotations for three tasks: semantic segmentation (13 classes), depth estimation, and surface normal prediction. Training uses standard task losses: per-pixel cross-entropy for segmentation, masked L1 for depth estimation over valid pixels, and a masked cosine loss over valid pixels for surface normal prediction. These choices match the NYU-v2 multi-task protocol followed by prior multi-task learning work~\citep{liu2023famo,navon2022multi}. Our implementation builds on the official Nash-MTL code base\footnote{\url{https://github.com/AvivNavon/nash-mtl}} and incorporates the FAMO implementation\footnote{\url{https://github.com/Cranial-XIX/FAMO}}. Following standard practice \citep{liu2019end, navon2022multi, liu2023famo, xiao2025ldc, shen2024go4align, dai2023improvable}, we evaluate segmentation using mean Intersection-over-Union (mIoU) and pixel accuracy; depth estimation using mean absolute error and mean relative error; and surface normals using median angular error, mean angular error, and the percentage of pixels whose angular error is below $30^\circ$, $22.5^\circ$, and $11.25^\circ$ \citep{silberman2012indoor}. These metrics provide a comprehensive view of per-task performance across classification, regression, and geometric prediction settings.

\paragraph{Model.} The model architecture used in our experiments is a Multi-Task Attention Network (MTAN)~\citep{liu2019end} built on top of SegNet~\citep{badrinarayanan2017segnet}. The network takes an RGB image as input and produces three task-specific outputs via separate heads: (i) a per-pixel class-score map for semantic segmentation, (ii) a scalar depth map for depth estimation, and (iii) a 3-channel surface-normal map for surface-normal prediction. For \texttt{LS}, \texttt{MGDA}, \texttt{UW}, \texttt{FAMO}, \texttt{NashMTL}, \texttt{IGBv2}, and \dibsmtl, we use the multi-task attention network (MTAN) \citep{liu2019end} with a SegNet-style encoder–decoder architecture and apply the same training schedule, optimizer, and data-augmentation procedures as established in prior work for a fair comparison. For \texttt{LDC} \citep{xiao2025ldc} and \texttt{GO4Align} \citep{shen2024go4align}, we use the task architectures provided in their original implementations.

\paragraph{Training Protocol.}For training, we follow the procedure outlined in~\citep{navon2022multi}. We train for 200 epochs with Adam~\citep{Kingma2014AdamAM}, using an initial learning rate of $1\times10^{-4}$ reduced to $5\times10^{-5}$ after 100 epochs. All reported results are averaged over three random seeds, namely $1, 7$, and $42$.

\begin{table}[t]
  \centering
  \caption{Hyperparameters used for MT10 (v1) SAC.}
  \label{tab:mt10-dibs-hparams}
  \setlength{\tabcolsep}{8pt}
  \begin{tabular}{l l}
    \toprule
    \textbf{Component} & \textbf{Value} \\
    \midrule
    Encoder feature dim & 50 \\
    Discount $\gamma$ & 0.99 \\
    Initial temperature $\alpha_0$ & 1.0 \\
    Actor update freq. & 1 step / env step \\
    Critic target $\tau$ & 0.005 \\
    Target update freq. & 1 step / env step \\
    Encoder EMA $\tau_{\text{enc}}$ & 0.05 \\
    Learning Rate & 0.025 \\
    Update-weights cadence & every step ($=1$) \\
    \bottomrule
  \end{tabular}
\end{table}

\subsection{Cityscapes}

\paragraph{Benchmark and Setup.} The \texttt{Cityscapes} outdoor driving scenes \citep{cordts2016cityscapes} provides 5000 RGB-D images with dense per-pixel annotations for 2 tasks: semantic segmentation (7 classes) and depth estimation. Training uses standard task losses: per-pixel cross-entropy for segmentation and masked L1 for depth estimation over valid pixels These choices match the Cityscapes multi-task protocol followed by prior multi-task learning work~\citep{liu2023famo,navon2022multi}.

\paragraph{Model.} Just as with NYU-v2, the model architecture used in our experiments is a Multi-Task Attention Network (MTAN)~\citep{liu2019end} built on top of SegNet~\citep{badrinarayanan2017segnet}. The network takes an RGB image as input and produces two task-specific outputs via separate heads: (i) a per-pixel class-score map for semantic segmentation, (ii) a scalar depth map for depth estimation. For \texttt{LS}, \texttt{MGDA}, \texttt{UW}, \texttt{FAMO}, \texttt{NashMTL}, \texttt{IGBv2}, and \dibsmtl, we use the multi-task attention network (MTAN) \citep{liu2019end} with a SegNet-style encoder–decoder architecture and apply the same training schedule, optimizer, and data-augmentation procedures as established in prior work for a fair comparison. For \texttt{LDC} \citep{xiao2025ldc} and \texttt{GO4Align} \citep{shen2024go4align}, we use the task architectures provided in their original implementations.

\paragraph{Training Protocol.} We follow the procedure outlined in~\citep{navon2022multi}. We train for 300 epochs with Adam~\citep{Kingma2014AdamAM}, using an initial learning rate of $1\times10^{-4}$ reduced to $5\times10^{-5}$ after 100 epochs. All reported results are averaged over three random seeds, namely $1, 7$, and $42$.

\subsection{QM9}
\paragraph{Benchmark and Setup.} The \texttt{QM9} dataset \cite{blum2009970} is a graph neural network learning becnhmark consisting of 11 tasks, and is a widely used challenging benchmark in MTL literature \citep{navon2022multi, liu2023famo, xiao2025ldc, dai2023improvable, shen2024go4align}. It consists
of molecules represented by annotated graphs. Our experimental setting is the same as that used in prior MTL works, and the tasks are to predict 11 different molecular properties.

\paragraph{Model.}
Following \citet{navon2022multi}, we use the PyTorch Geometric implementation of \citet{fey2019pytorch}, which incorporates the widely used GNN architecture of \citet{gilmer2017neural} together with the pooling operator of \citet{vinyals2015order}. As with the other benchmarks, \texttt{LDC} and \texttt{GO4Align} use their own architectures.

\paragraph{Training Protocol.}
We train each method for 300 epochs and perform a learning-rate search based on the validation $\Delta m$ metric. A learning-rate scheduler reduces the learning rate once the validation $\Delta m$ stops improving, and the validation set is additionally used for early stopping. All reported results are averaged over three random seeds, namely 1, 7,
and 42. This is in line with the approach taken by ~\cite{navon2022multi}.

\subsection{Meta-World MT10}

\paragraph{Benchmark and Setup.}
The MetaWorld MT10 (v1) benchmark comprises ten robotic manipulation tasks: \texttt{Reach}, \texttt{Push}, \texttt{Pick-and-Place}, \texttt{Door Open}, \texttt{Drawer Open}, \texttt{Drawer Close}, \texttt{Button Press (Top-Down)}, \texttt{Peg Insert }, \texttt{Window Open}, and \texttt{Window Close}~\citep{yu2020meta}. Each task has distinct reward functions and success criteria. We build on the MTRL code base~\citep{Sodhani2021MTRL}\footnote{\url{https://github.com/facebookresearch/mtrl}} with MetaWorld\footnote{\url{https://github.com/Farama-Foundation/Metaworld}}.

\paragraph{Policy and Training Protocol.}
We train a single Soft Actor–Critic (SAC) policy shared across all ten tasks. multi-task learning (MTL) methods are applied to balance the actor and critic updates within the shared SAC. For each MTL method, we train for $2{,}000{,}000$ environment steps in total. Episodes have length $150$; this corresponds to $\approx 13{,}333$ episodes overall. Evaluation is performed every $200$ episodes ($30{,}000$ steps) on all tasks, and reported metrics are averaged over 10 random seeds: $1,2,3,4,5,6,7,8,9,10$.

\begin{figure}[t]
  \centering

  \begin{subfigure}[t]{0.48\linewidth}
    \centering
    \begin{tikzpicture}
\path[use as bounding box] (0,0) rectangle (12.5cm,7.8cm);
\begin{axis}[
    ybar,
    xshift=13mm,
    ymin=0, ymax=0.011,
    ymajorgrids,
    bar width=30pt,
    width=\linewidth, height=6.5cm,
    enlarge x limits=0.35,
    x=3.2cm,
    symbolic x coords={1-Step Nash-MTL,1-Step DiBS-MTL},
    xtick=data,
    xticklabels={\texttt{Nash-MTL},\texttt{1-Step DiBS-MTL}},
    ylabel={Seconds Per Iteration},
    tick align=outside,
    tick style={semithick},
    nodes near coords,
    every node near coord/.append style={
        /pgf/number format/sci,
        /pgf/number format/sci e,
        /pgf/number format/precision=2
    },
    font=\small
]

\addplot+[]
coordinates {
  (1-Step Nash-MTL,0.00923)
  (1-Step DiBS-MTL,0.000915)
};

\end{axis}
\end{tikzpicture}
    \caption{Average seconds per iteration ($\downarrow$ is faster) in the nonconvex example from \Cref{sec:toy}.}
  \end{subfigure}
  \hfill
  \begin{subfigure}[t]{0.48\linewidth}
    \centering
    \begin{tikzpicture}
\path[use as bounding box] (0,0) rectangle (12.5cm,7.8cm);
\begin{axis}[
    ybar,
    ymin=0, ymax=70,
    ymajorgrids,
    bar width=30pt,
    width=\linewidth, height=6.5cm,
    enlarge x limits=0.15,
    x=3.2cm,
    symbolic x coords={1-Step Nash-MTL,100-Step Nash-MTL,1-Step DiBS-MTL},
    xtick=data,
    xticklabels={
        \texttt{Nash-MTL},
        \texttt{Nash-MTL} 1 in 100 steps,
        \texttt{1-Step DiBS-MTL}
    },
    ylabel={Seconds per epoch},
    tick align=outside,
    tick style={semithick},
    nodes near coords,
    every node near coord/.append style={/pgf/number format/fixed},
    font=\small
]

\addplot+[]
coordinates {
  (1-Step Nash-MTL,49.940)
  (100-Step Nash-MTL,8.214)
  (1-Step DiBS-MTL,6.331)
};

\end{axis}
\end{tikzpicture}
    \caption{Average seconds per epoch ($\downarrow$ is faster) for the multi-task RL example in \Cref{subsection: mtrl}.}
  \end{subfigure}

  \caption{Runtime comparisons across nonconvex demonstrative example and multi-task RL settings.}
  \label{fig:runtime-comparison}

  \vspace{1em}

  \captionof{table}{Run times in minutes ($\downarrow$ is faster) for supervised MTL experiments in \Cref{subsection: cv/mtrl} averaged over 3 seeds.}
  \label{tab:runtime_mtl}

  \setlength{\tabcolsep}{8pt}
  \begin{tabular}{lccc}
  \toprule
   & \multicolumn{3}{c}{\textbf{Benchmarks}} \\
  \cmidrule(lr){2-4}
   & \texttt{NYU-V2} & \texttt{Cityscapes} & \texttt{QM9} \\
    \midrule
    \texttt{1-step DiBS-MTL} & \textbf{819.33} & \textbf{481.67} & 1705 \\
    \texttt{5-step DiBS-MTL} & 1713 & 612 & 1719 \\
    \texttt{10-step DiBS-MTL} & 1799 & 752 & 1726 \\
    \texttt{Nash-MTL} & 1031 & 598 & \textbf{1699} \\
    \bottomrule
  \end{tabular}

\end{figure}

\paragraph{Hyperparameters.}
Key hyperparameter values are provided in \Cref{tab:mt10-dibs-hparams}. We follow the same hyperparemeters defined in prior works ~\citep{navon2022multi,liu2023famo,liu2021conflict,ban2024fair}. 

\paragraph{Reproducibility Note.}
During preliminary experiments, we observed that using the same random seed did not always produce identical training curves. This discrepancy arose because the original MTRL code \citep{Sodhani2021MTRL} did not seed the initial action sampling during the exploration phase. As a result, small differences in early exploratory actions (first 10 steps) propagated through training and produced slight variation in learning behavior even with fixed seeds.

\section{Additional Results}\label{section: additional results}
\subsection{Comparison of \dibsmtl~ runtimes with previous bargaining MTL method \texttt{Nash-MTL}.}\label{subsection: runtime}
We also evaluated the computational speed of the \dibsmtl~with the previous bargaining-inspired MTL method, \texttt{Nash-MTL}. Specifically, we compare runtimes of \dibsmtl~ with Nash-MTL for the demonstrative nonconvex examples (\Cref{fig:runtime-comparison}), the multi-task reinforcement learning example (\Cref{fig:runtime-comparison}), and the supervised learning \texttt{NYU-v2}, \texttt{QM9} and \texttt{Cityscapes} examples (\Cref{tab:runtime_mtl}).

From \Cref{fig:runtime-comparison}, it is clear that \dibsmtl~consistently is faster than \texttt{Nash-MTL} for the demonstrative nonconvex and the multi-task reinforcement learning examples. In the multi-task reinforcement example particularly, \texttt{Nash-MTl} is particularly slow if applied for every training iteration, and even when applied only once per 100 training loops (which is the speedup suggested by the original authors \citep{navon2022multi}), it is still slower than \ref{eq: dibs mtl}. We conjecture that the slow performance of \texttt{Nash-MTL} is because it solves a separate convex optimization problem at each iteration which is time consuming. 

In \Cref{tab:runtime_mtl}, we compare the runtime of different \dibsmtl~ variants and \texttt{Nash-MTL} across the MTL benchmarks to assess how the number of bargaining steps affects computational cost. The single-step version of \dibsmtl~ achieves the fastest runtimes on \texttt{NYU-V2} and \texttt{Cityscapes}, while falling slightly behind \texttt{Nash-MTL} on \texttt{QM9}. As expected, runtime for \dibsmtl~variants increases as the number of steps increase, because the per epoch complexity increases, reflecting the added cost of performing multiple bargaining updates. At the same time, as indicated by \Cref{tab:nyu,tab:qm9,tab:cityscapes}, \ref{eq: dibs mtl} achieves comparable performance to the multi-step versions, making \ref{eq: dibs mtl} an appropriate approximation of multi-step \dibsmtl~variants.

\begin{figure}[t]
  \centering

  \begin{subfigure}[t]{\textwidth}
    \centering
    \begin{subfigure}[t]{0.19\textwidth}\panel{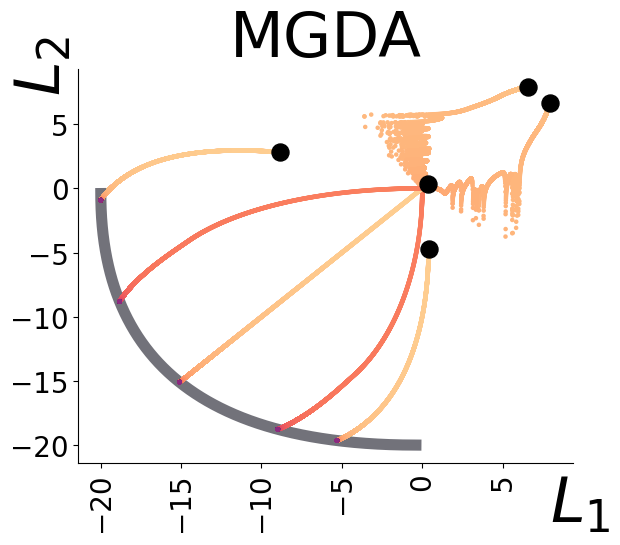}\end{subfigure}\hfill
    \begin{subfigure}[t]{0.19\textwidth}\panel{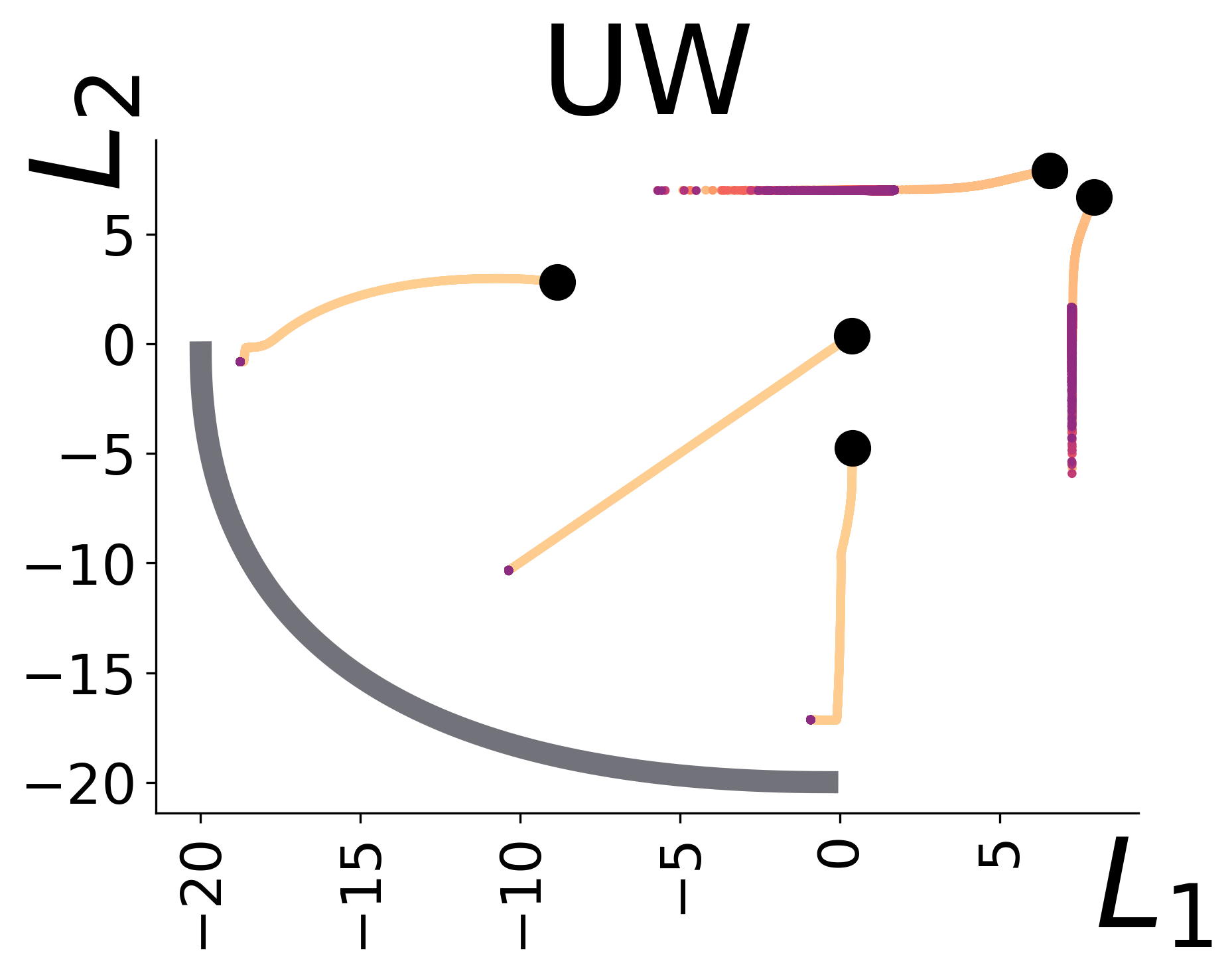}\end{subfigure}\hfill
    \begin{subfigure}[t]{0.19\textwidth}\panel{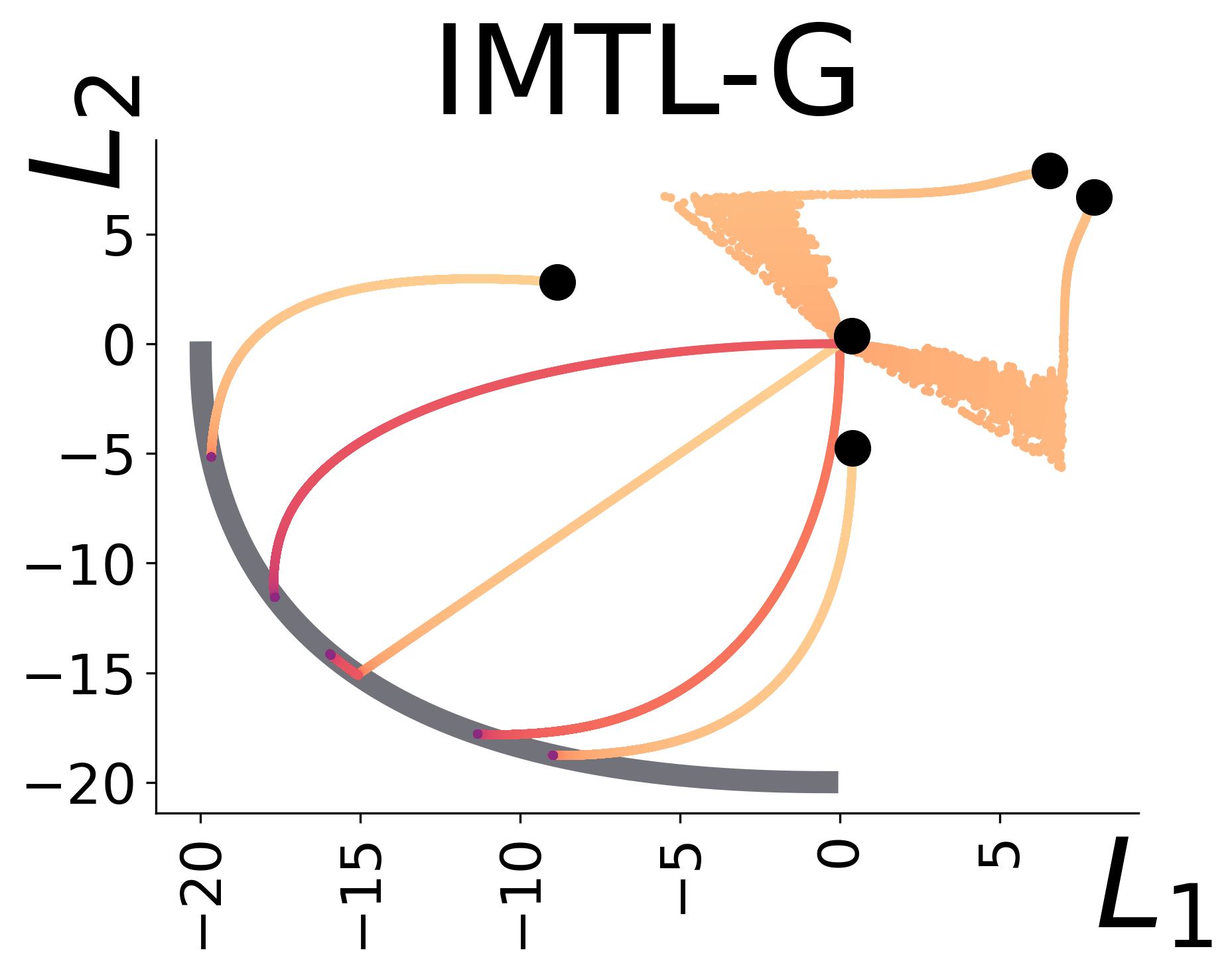}\end{subfigure}\hfill
    \begin{subfigure}[t]{0.19\textwidth}\panel{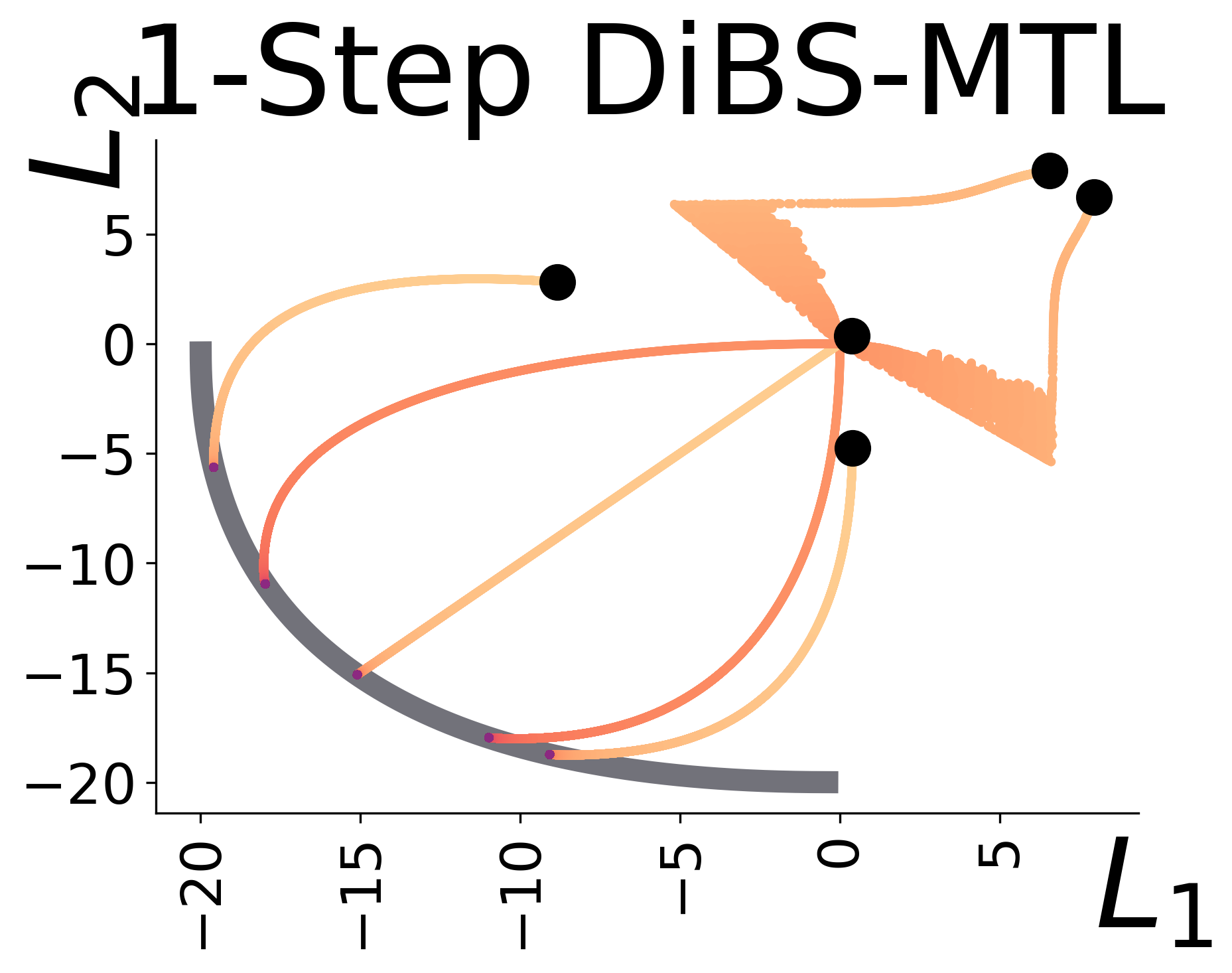}\end{subfigure}
    \subcaption*{Untransformed}
  \end{subfigure}

  \vspace{0.9em}

  \begin{subfigure}[t]{\textwidth}
    \centering
    \begin{subfigure}[t]{0.19\textwidth}\panel{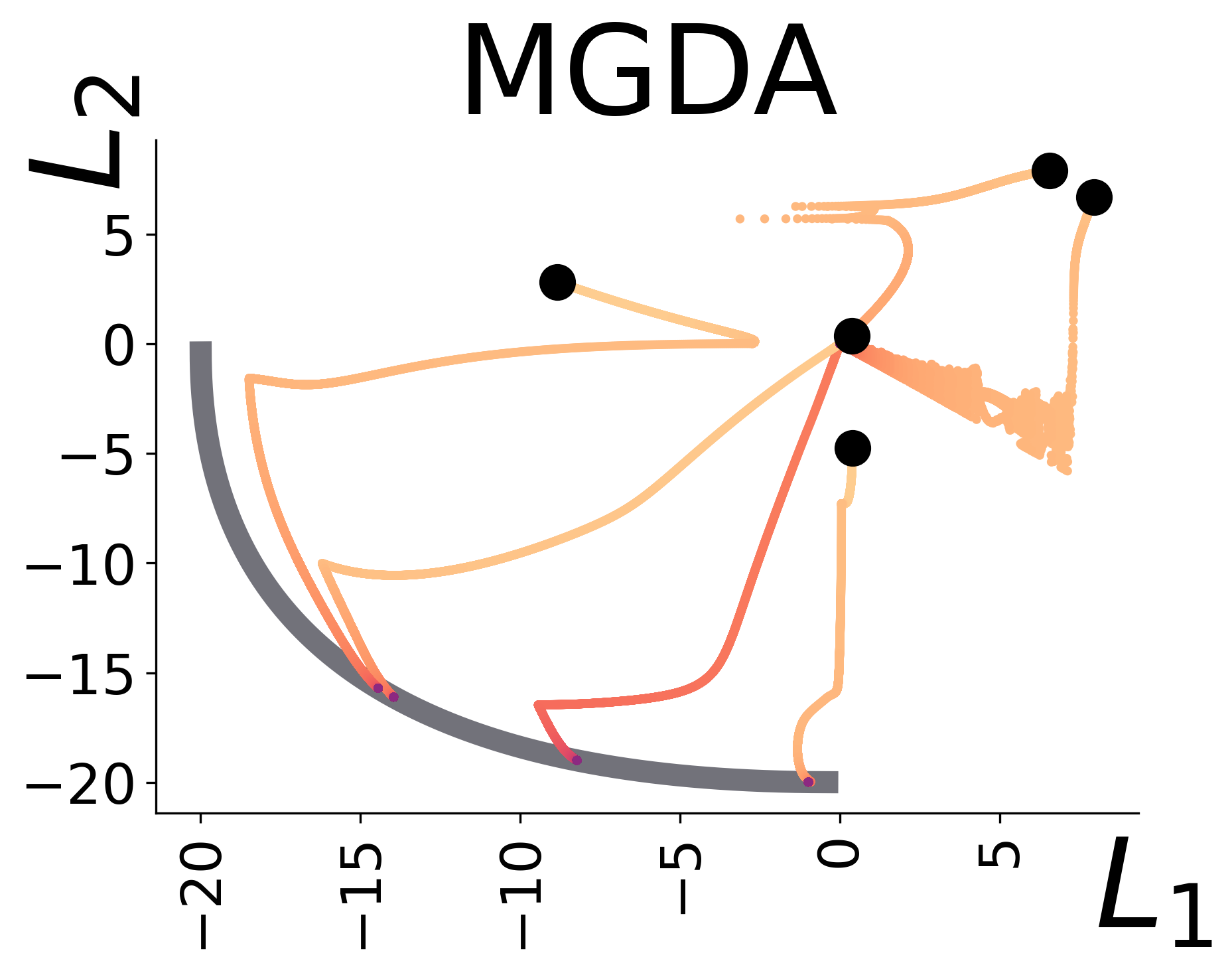}\end{subfigure}\hfill
    \begin{subfigure}[t]{0.19\textwidth}\panel{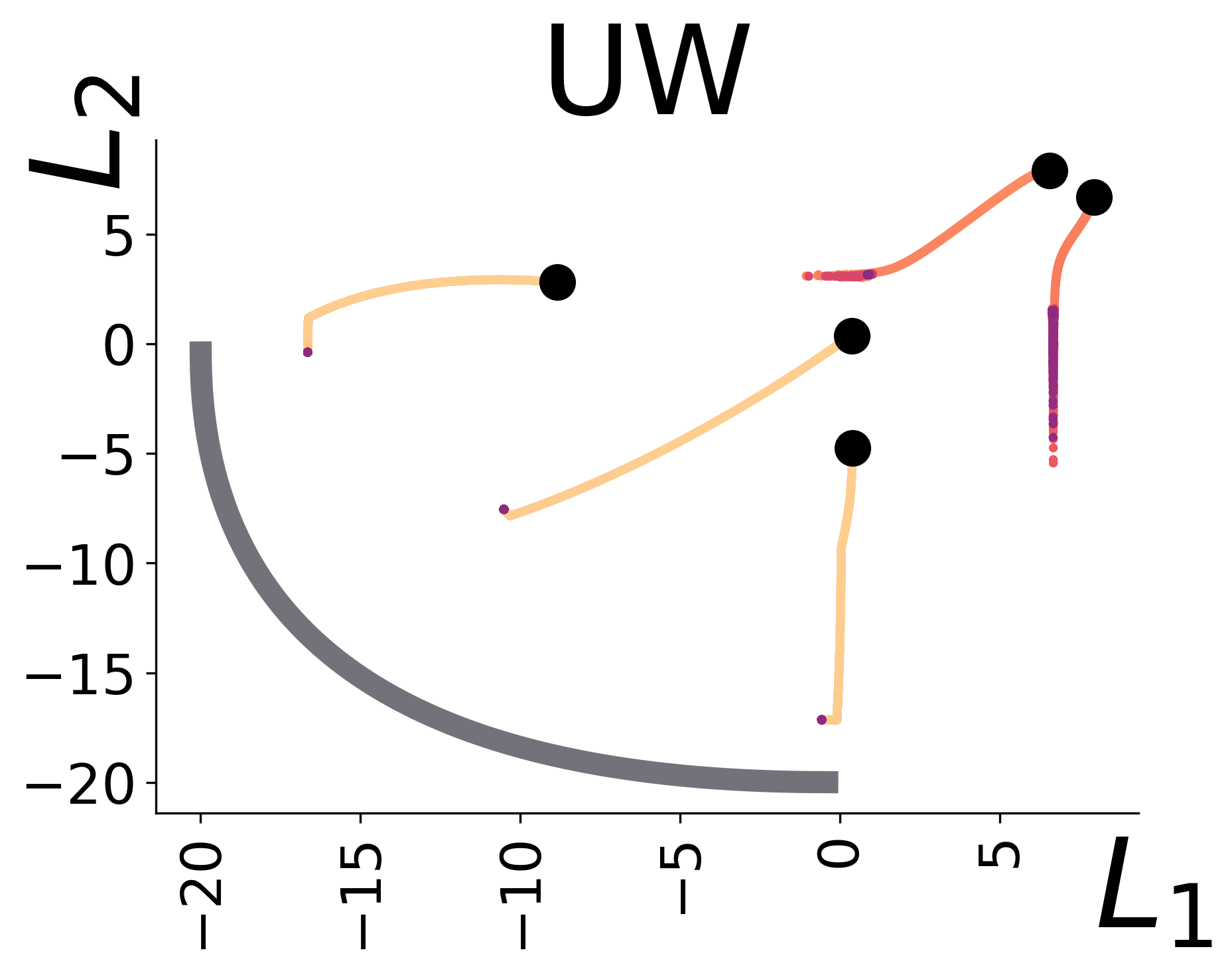}\end{subfigure}\hfill
    \begin{subfigure}[t]{0.19\textwidth}\panel{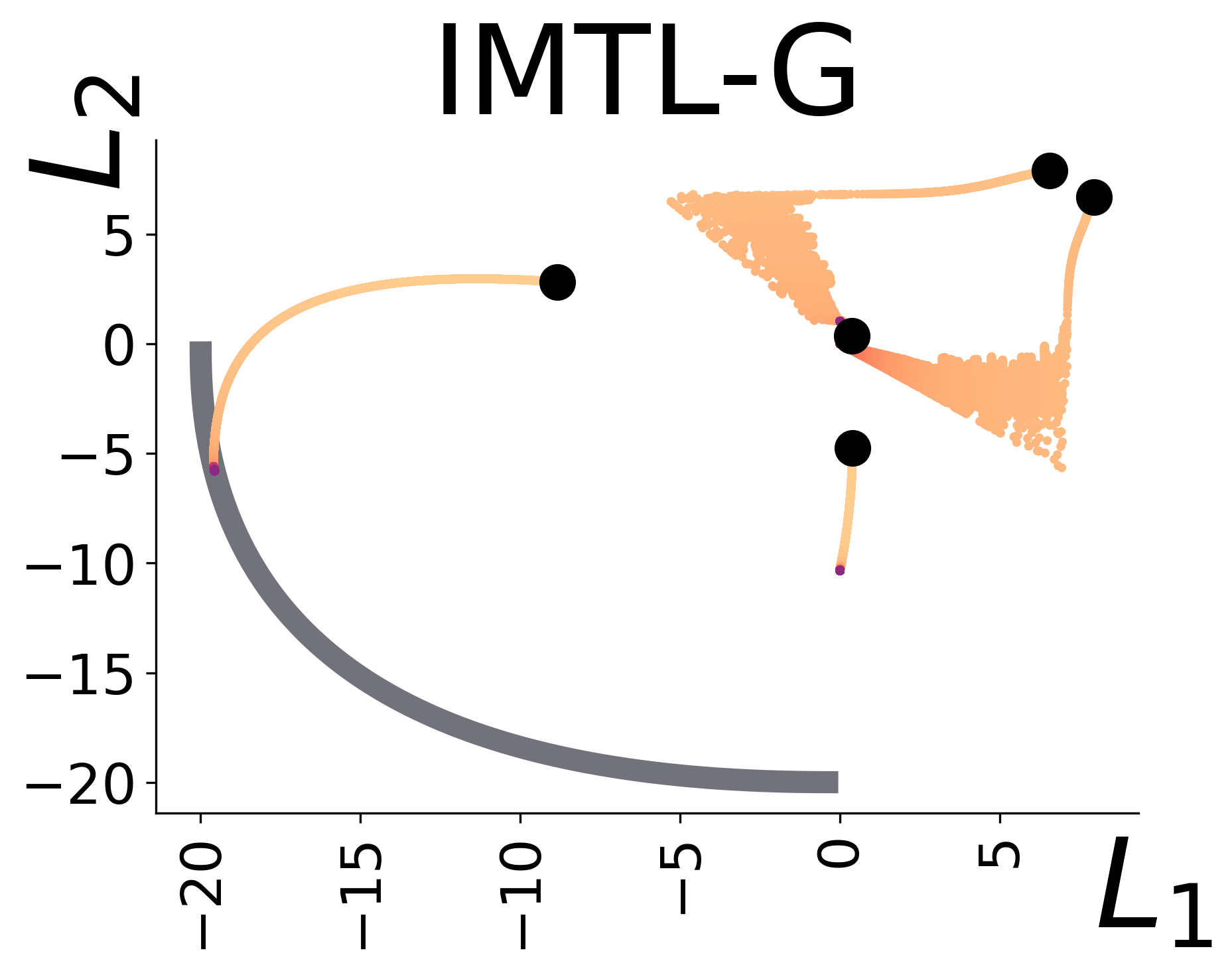}\end{subfigure}\hfill
    \begin{subfigure}[t]{0.19\textwidth}\panel{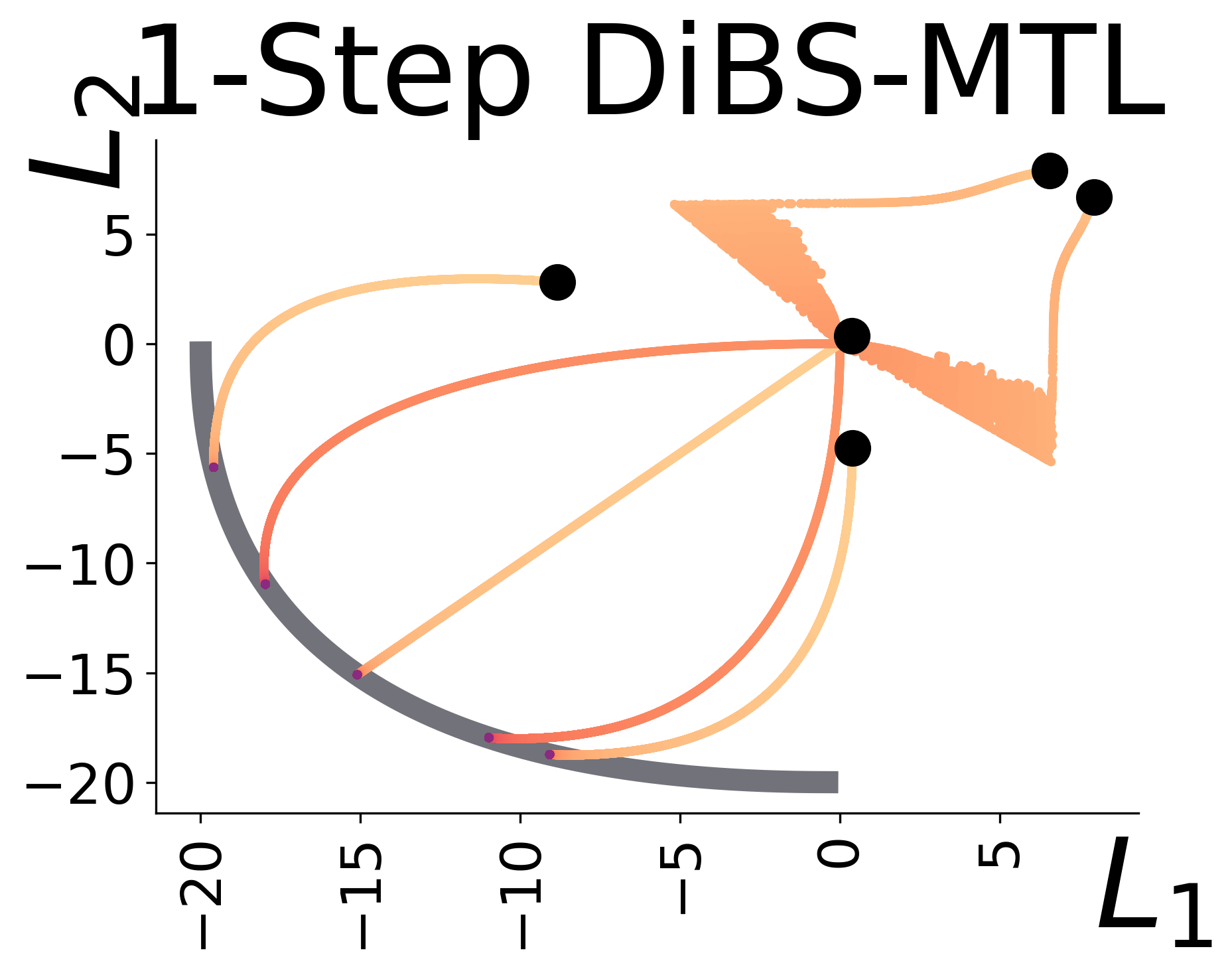}\end{subfigure}
    \subcaption*{Transformed}
  \end{subfigure}

  \caption{Additional experiments performed in the demonstrative two-objective example. We observe that, \ref{eq: dibs mtl}~exhibits the same invariance to monotone non-affine transforms as \texttt{T-step} \dibsmtl ~from \Cref{fig:trajectories}, and also demonstrates similar behaviour, tracing very similar trajectories as  \texttt{T-step} \dibsmtl ~($T=5$) from the same initializations.}
  \label{fig:extra_trajectories}
\end{figure}

\begin{algorithm}[t]
\caption{\dibsmtl\ Single Step}
\label{alg:dibs-mtl-ss}
\begin{algorithmic}

\REQUIRE Initial parameters $\theta_{(0)}$, losses $\{\ell^i\}_{i=1}^N$, 
learning rate $\eta$, number of epochs $J$

\FOR{$j = 1$ to $J$}

    \STATE Compute task gradients 
    $\nabla_{\theta_{(j-1)}} \ell^i$

    \STATE Compute \texttt{DiBS} update direction:
    
    \(
    d_{(j)} =
    \sum_{i\in[N]}
    \normalizedgradi{\theta}{\theta_{(j-1)}}
    \)

    \STATE Update:
    \(
    \theta_{(j)} \leftarrow
    \theta_{(j-1)} - \eta\, d_{(j)}
    \)

\ENDFOR

\RETURN $\theta_{(J)}$

\end{algorithmic}
\end{algorithm}

\subsection{Additional Results in the Demonstrative Nonconvex Example}

In addition to the results reported in section \Cref{sec:toy}, we also ran additional methods in the demonstrative two-objective example. Specifically, we ran the Multiple Gradient Descent Algorithm (\texttt{MGDA}) ~\citep{desideri2012multiple}, Uncertainty Weighting (\texttt{UW}) ~\citep{kendall2018multi}, Impartial Multi-task Learning (IMTL-G) ~\citep{liu2021towards}, and \ref{eq: multi step dibs}.  We run \ref{eq: multi step dibs} for 10 steps per update. The loss functions and number of steps are identical to the experiment described in \Cref{sec:toy}.

In \Cref{fig:extra_trajectories}, we observe that the additional baseline methods of MGDA, UW, IMTL-G are not invariant to the transform, with UW failing to reach the Pareto-front in both cases. We note that \ref{eq: multi step dibs} performs similarly to single step \dibsmtl~. \ref{eq: multi step dibs} reaches the same points on the Pareto-front in both cases, showing it is also invariant to monotone non-affine transforms.

\section{\texttt{1-step DiBS-MTL} Algorithm Pseudocode}

 \Cref{alg:dibs-mtl-ss} provides pseudocode for the \ref{eq: dibs mtl} algorithm. When adapting \dibsmtl~ to the single-step variant, the distances to the preferred states for each agent are equal. For \ref{eq: dibs mtl}, for all $j$ in \Cref{alg:dibs-mtl}, $\|\theta_{(j-1)} - \theta^{*,i}\| = \epsilon$ for all $i$. \textit{The approximation radius $\epsilon$ and the step size $\eta_s$ are absorbed into the learning rate $\eta$.}

\end{document}